\documentclass[sigconf]{acmart}

\AtBeginDocument{%
}

\usepackage{mathtools}
\usepackage{subcaption}
\usepackage{multirow}
\usepackage{algorithm}
\usepackage{algorithmic}
\usepackage{colortbl}
\usepackage[capitalize,noabbrev]{cleveref}

\newcommand{\boldres}[1]{\textbf{#1}}
\newcommand{\secondres}[1]{\underline{#1}}
\definecolor{tpsblue}{RGB}{225,240,250}

\theoremstyle{definition}

\theoremstyle{remark}

\copyrightyear{2026}
\acmYear{2026}
\setcopyright{cc}
\setcctype{by}

\acmConference[CIKM '26]
{Proceedings of the 35th ACM International Conference on Information and Knowledge Management}
{November 07--11, 2026}
{Rome, Italy}

\acmBooktitle{Proceedings of the 35th ACM International Conference on Information and Knowledge Management (CIKM '26), November 07--11, 2026, Rome, Italy}

\acmDOI{10.1145/3799682.3840727}
\acmISBN{979-8-4007-2539-5/2026/11}

\begin{document}


\title{Sliding-Window Reordering with Overlap Averaging:
A Simple Time-Domain Augmentation for Multivariate Forecasting}


\author{Jafar Bakhshaliyev}
\correspondingauthor
\orcid{0009-0008-7846-5680}

\affiliation{%
  \department{Data Science Group}
  \institution{University of Hildesheim}
  \city{Hildesheim}
  \country{Germany}
}

\email{bakhshaliyevj@uni-hildesheim.de}

\author{Johannes Burchert}
\orcid{0009-0002-6832-977X} 

\affiliation{%
  \department{Information Systems and Machine Learning Lab (ISMLL)}
  \institution{University of Hildesheim}
  \city{Hildesheim}
  \country{Germany}
}

\email{burchert@ismll.uni-hildesheim.de}

\author{Niels Landwehr}
\orcid{0000-0001-5019-7620}

\affiliation{%
  \department{Data Science Group}
  \institution{University of Hildesheim}
  \city{Hildesheim}
  \country{Germany}
}

\email{landwehr@uni-hildesheim.de}

\author{Lars Schmidt-Thieme}
\orcid{0000-0001-5729-6023}

\affiliation{%
  \department{Information Systems and Machine Learning Lab (ISMLL)}
  \institution{University of Hildesheim}
  \city{Hildesheim}
  \country{Germany}
}

\email{schmidt-thieme@ismll.de}

\renewcommand{\shortauthors}{Bakhshaliyev et al.}


\begin{abstract}

Augmentation has become a central technique for improving deep forecasting models, but classification-style transformations tend to break the coherence between the look-back window and its continuous future target. We describe a simple procedure that unfolds the joint input--target sequence into overlapping sliding windows, randomly reorders a controlled fraction of them---prioritized by a lightweight variance criterion---and reconstructs the sequence by averaging across the overlaps, producing synthetic samples with controlled variation while limiting temporal distortion. The procedure is model-agnostic, introduces only three interpretable hyperparameters, and achieves strong improvements over a comprehensive set of competing augmentations across nine long-term forecasting benchmarks with five backbone families (TSMixer, DLinear, PatchTST, TiDE, LightTS) and four short-term traffic benchmarks with PatchTST. Component-wise ablations, hyperparameter sensitivity studies, distributional-alignment diagnostics, probabilistic forecasting evaluation, and a transfer experiment to univariate and multivariate time series classification clarify the contribution of each design choice.

\end{abstract}





\maketitle


\section{Introduction}

Multivariate time series prediction forecasts future values for multiple interdependent variables or channels. Time series forecasting is crucial in many areas, such as finance, healthcare, meteorology, and manufacturing~\cite{zhou2021informerefficienttransformerlong, zeng2022transformerseffectivetimeseries, liu2024itransformerinvertedtransformerseffective}. However, in many real-world applications, the available sensor data is limited and the underlying temporal patterns depend on additional external factors that are not directly observable~\cite{chen2023fraugfrequencydomainaugmentation, upsample, zhao2024dominantshufflesimplepowerful}.

Data augmentation is becoming increasingly important in time series forecasting (TSF), as it plays a key role in boosting model accuracy and strengthening generalization capabilities. When real data are scarce or insufficient, synthetic sample generation becomes essential. Augmentation enriches the dataset by applying transformations or perturbations to existing sequences, broadening the diversity of temporal patterns and improving model robustness against noise~\cite{Wen_2021, chen2023fraugfrequencydomainaugmentation, arabi2024wavemaskmixexploringwaveletbasedaugmentations, zhao2024dominantshufflesimplepowerful}.

Although many augmentation methods have been proposed for TSF, designing an effective technique that preserves temporal dynamics and coherence remains challenging~\cite{chen2023fraugfrequencydomainaugmentation, zhao2024dominantshufflesimplepowerful}. Unlike classification, forecasting requires preserving coherence between the input window and its continuous future target. Transformation-based methods, such as noise injection, scaling, or time warping, are effective for time series classification but generally fail to deliver substantial benefits in forecasting~\cite{leguennec:halshs-01357973, Um_2017, Wen_2021, chen2023fraugfrequencydomainaugmentation}. Recent research has therefore focused on frequency-based augmentation, which manipulates the spectral components of time series and currently represents the most competitive family of augmentation methods for TSF~\cite{chen2023fraugfrequencydomainaugmentation, arabi2024wavemaskmixexploringwaveletbasedaugmentations, zhao2024dominantshufflesimplepowerful}. Decomposition-based and other augmentation techniques~\cite{mbb, upsample, zhang2023diversecoherentaugmentationtimeseries} have also shown promise, though comprehensive comparisons under unified fair experimental settings remain limited.

In computer vision (CV), patch-based augmentations such as PatchShuffle~\cite{kang2017patchshuffleregularization} and PatchMix~\cite{hong2024patchmix} have proven highly effective. However, to the best of our knowledge, no patch-based augmentation method has been developed specifically for TSF tasks. A naive adaptation of such image-style techniques to temporal data fails because it destroys local temporal coherence, often introducing artifacts and leading to distribution shifts. To address this gap, we propose Temporal Patch Shuffle (TPS), a forecasting-tailored augmentation method that operates on overlapping temporal patches, applies controlled shuffling, and reconstructs the sequence by averaging overlapping regions. TPS is designed to introduce structured variation while preserving forecast-consistent local temporal structure and limiting the distributional and temporal distortions induced by patch reordering.

Our contributions are threefold:
\begin{itemize}
    \item We propose \textbf{Temporal Patch Shuffle (TPS)}, a simple, model-agnostic augmentation method for forecasting that extracts \emph{overlapping} temporal patches, applies controlled shuffling with variance-based ordering as a conservative heuristic, and reconstructs the sequence by averaging overlaps.

    \item We demonstrate TPS on both long-term and short-term forecasting: across nine long-term benchmarks and five backbones, TPS improves average MSE by \textbf{2.08--10.51\%} relative to the strongest competing augmentation; on four short-term traffic benchmarks with PatchTST, TPS obtains the best or tied-best average MSE on all four datasets and the best average MAE on all four, with MSE reductions of up to \textbf{7.14\%}.

    \item We provide extensive analyses, including component-wise ablations, hyperparameter sensitivity, distributional-alignment diagnostics, probabilistic forecasting quality, and computational-overhead analysis, and we further show that TPS transfers beyond forecasting to time series classification.

\end{itemize}

The code for this work is available at: \url{https://github.com/jafarbakhshaliyev/TPS}. An extended version with per-dataset results, full per-length tables, additional ablations (augmentation size and ratio), and classification experimental details is provided in the supplementary material: \url{https://github.com/jafarbakhshaliyev/TPS/blob/main/CIKM2026_TPS_Supplementary_Material.pdf}.

\begin{figure*}[ht]
\begin{center}
    \centerline{\includegraphics[width=\textwidth]{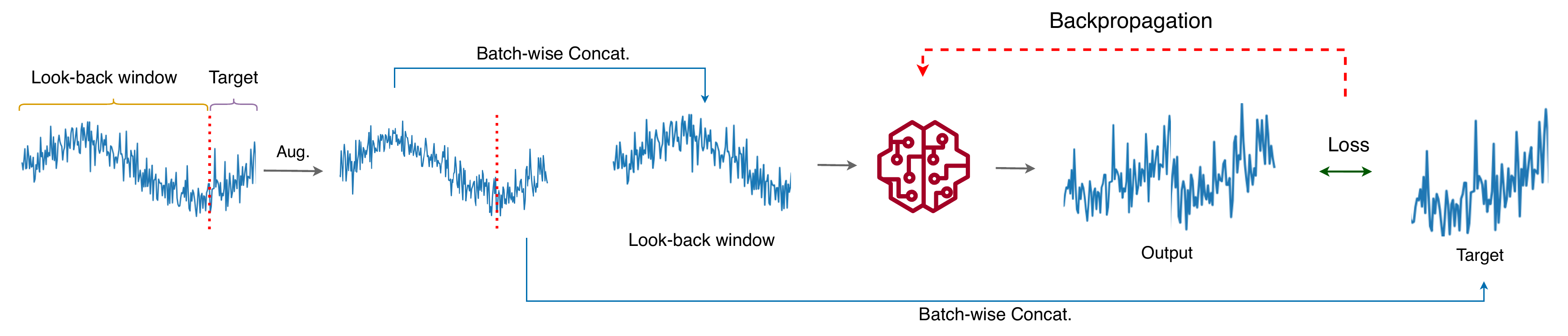}}
    \caption{
    Overview of the training pipeline for time series forecasting with augmentation.
    The look-back window and forecast horizon are concatenated and processed by the augmentation module to produce synthetic sequences, following the general procedure described in \citet{chen2023fraugfrequencydomainaugmentation}.
    }
    \Description{
Training pipeline for time series forecasting with augmentation.
The look-back window and target horizon are jointly augmented,
the original and augmented samples are concatenated along the batch
dimension, and a forecasting model produces an output that is compared
with the corresponding target to compute the loss and backpropagate gradients.
}
    \label{fig:tsf-frame}
\end{center}
\end{figure*}

\section{Related Work}

A wide range of data augmentation methods have been proposed to improve the generalization and robustness of TSF models. Existing approaches can be grouped into four major categories: transformation-based methods~\cite{cui2016multiscaleconvolutionalneuralnetworks,wen2019timeseriesanomalydetection,Wen_2021}, frequency-based augmentations~\cite{gao2021robusttadrobusttimeseries,chen2023fraugfrequencydomainaugmentation,freqpool,arabi2024wavemaskmixexploringwaveletbasedaugmentations, zhao2024dominantshufflesimplepowerful}, decomposition-based techniques~\cite{zhang2023diversecoherentaugmentationtimeseries}, and other augmentation methods~\cite{asd,timegan,mbb,upsample}. To the best of our knowledge, patch-based augmentation methods tailored specifically to time series forecasting have not been explored.

\paragraph{Transformation-based Augmentations.}
Initially developed in computer vision, transformation-based methods such as Gaussian noise injection~\cite{wen2019timeseriesanomalydetection}, window cropping~\cite{cui2016multiscaleconvolutionalneuralnetworks}, window warping, and other techniques~\cite{Wen_2021} have been adapted for time series tasks including classification and anomaly detection, showing significant effectiveness. Nevertheless, the direct application of these augmentation techniques to TSF tasks generally does not produce substantial benefits, as they either disturb temporal order, for example by introducing random noise or shifting the time series, or lack sufficient diversity in the generated augmented samples, both of which are important considerations for forecasting tasks~\cite{Wen_2021,zhang2023diversecoherentaugmentationtimeseries,zhao2024dominantshufflesimplepowerful}. \citet{chen2023fraugfrequencydomainaugmentation} evaluated these techniques and concluded that they generally do not improve performance over models trained without augmentation.

\paragraph{Frequency-based Augmentations.}
Using time-frequency information, several recent works have introduced augmentation strategies designed to improve forecasting performance, contributing to an expanding set of frequency-based techniques. One of the earliest approaches, RobustTAD~\cite{gao2021robusttadrobusttimeseries}, perturbs either the magnitude or phase of the Fourier spectrum. \citet{chen2023fraugfrequencydomainaugmentation} proposed two augmentation techniques: \emph{FreqMask}, which masks the signal's frequency components, thereby removing specific events from the underlying system, and \emph{FreqMix}, which mixes frequencies from two random series to exchange structural behaviors between systems. Wavelet-based extensions, including \emph{WaveMask} and \emph{WaveMix}~\cite{arabi2024wavemaskmixexploringwaveletbasedaugmentations}, address a limitation of Fourier-based methods by operating in a joint time--frequency space, enabling multi-resolution and localized perturbations. More recently, \citet{zhao2024dominantshufflesimplepowerful} proposed \emph{Dominant Shuffle}, which shuffles the top dominant frequency components and mitigates the out-of-distribution artifacts produced by FreqMask and FreqMix. Additional frequency-domain methods include \emph{FreqAdd}~\cite{freqadd}, which perturbs targeted frequency components by additive modification, and \emph{FreqPool}~\cite{freqpool}, which compresses the spectrum via pooling operations to improve model robustness.

\paragraph{Decomposition-based Augmentations and Others.}
Several TSF augmentation strategies build on decomposition methods, including Seasonal and Trend decomposition using Loess (STL)~\cite{cleveland1990stl} and Empirical Mode Decomposition (EMD)~\cite{huang1998empirical}. Early work by \citet{nam2020data} applied EMD to decompose a series into Intrinsic Mode Functions (IMFs), which capture oscillatory components from high-frequency fluctuations to low-frequency variations, and generated augmentations by filtering noise. More recently, Spectral and Time Augmentation (STAug)~\cite{zhang2023diversecoherentaugmentationtimeseries} introduced a strategy that selects two sequences, applies EMD decomposition, weights IMF components, and combines them using a mixup-style interpolation.

Beyond decomposition-based methods, several other augmentation methods have been proposed. The Upsample technique~\cite{upsample} selects consecutive segments and expands them back to the original length using linear interpolation, effectively acting as a ``magnifying glass'' that emphasizes local patterns. Weighted Dynamic Time Warping Barycentric Averaging (wDBA)~\cite{asd} generates augmented samples by computing DTW distances and applying an exponentially weighted average over the closest neighbors in the training set. Moving Block Bootstrapping (MBB)~\cite{mbb} decomposes a time series into trend, seasonal, and remainder components via STL, and perturbs the remainder using bootstrapped blocks to produce new time series.

\section{Problem Formulation}

We focus on the multivariate time series forecasting (MTSF) setting, where the goal is to predict future values for multiple correlated variables or channels over time.
Forecasting tasks are commonly categorized by prediction length into short-term and long-term settings; longer prediction lengths introduce more uncertainty and are therefore more challenging~\cite{zhao2024dominantshufflesimplepowerful}.

We use 0-based indexing and Python-style slicing notation throughout, where $a\!:\!b$ denotes the half-open interval $[a,b)$.

A multivariate time series batch of size $B$, length $T$, with $C$ channels is represented as
\begin{equation}
\mathbf{X} \in \mathbb{R}^{B \times T \times C}.
\end{equation}
We denote by $\mathbf{X}[b,\tau,:]\in\mathbb{R}^{C}$ the observation vector at time step $\tau$ for the $b$-th instance.

Given a look-back window length $t<T$, we define the look-back window as
\begin{equation}
\mathbf{L} = \mathbf{X}[:,\,0:t,\,:] \in \mathbb{R}^{B \times t \times C},
\end{equation}
and the corresponding forecast horizon of length $h=T-t$ as
\begin{equation}
\mathbf{F} = \mathbf{X}[:,\,t:T,\,:] \in \mathbb{R}^{B \times h \times C}.
\end{equation}

A forecasting model $f_{\boldsymbol{\theta}}$ parameterized by $\boldsymbol{\theta}$ learns the mapping
\begin{equation}
f_{\boldsymbol{\theta}}: \mathbb{R}^{B \times t \times C} \rightarrow \mathbb{R}^{B \times h \times C},
\quad
\mathbf{L} \mapsto f_{\boldsymbol{\theta}}(\mathbf{L}).
\end{equation}

Model performance is evaluated using the Mean Squared Error (MSE):
\begin{equation}
\label{eq:mse}
\text{MSE} = \frac{1}{B \cdot h \cdot C}\left\| \mathbf{F} - f_{\boldsymbol{\theta}}(\mathbf{L}) \right\|_{F}^{2},
\end{equation}
where $\|\cdot\|_F$ denotes the Frobenius norm.

Data augmentation methods aim to improve generalization by enriching the training distribution. In forecasting, however, augmentation must preserve coherence between the look-back window and its continuous future target, rather than perturbing the input in isolation. Augmentations therefore generate synthetic sequences $\mathbf{S}$ that are later split into synthetic look-back windows and corresponding synthetic forecast horizons.

Figure~\ref{fig:tsf-frame} provides an overview of the forecasting augmentation pipeline. Before augmentation is applied, the look-back window and its associated forecast horizon are concatenated to preserve input--target alignment. Augmentation produces synthetic look-back windows $\mathbf{S}_L$ and corresponding synthetic forecast horizons $\mathbf{S}_F$. We then form the augmented look-back window $\overline{\mathbf{L}}$ and augmented forecast horizon $\overline{\mathbf{F}}$ by concatenating synthetic and original samples along the batch dimension, yielding a batch of size $2B$:
\begin{equation}\label{eq:concat}
\begin{aligned}
\overline{\mathbf{L}} &= [ \mathbf{L}; \mathbf{S}_L ] \in \mathbb{R}^{2B \times t \times C},\\
\overline{\mathbf{F}} &= [ \mathbf{F}; \mathbf{S}_F ] \in \mathbb{R}^{2B \times h \times C}.
\end{aligned}
\end{equation}
The forecasting model is trained on $(\overline{\mathbf{L}}, \overline{\mathbf{F}})$.

\section{Motivation}

While many time series augmentation techniques are inspired by advances in computer vision (CV), patch-based augmentations, widely used in CV, remain largely unexplored in time series analysis. Two recent CV methods, PatchShuffle~\cite{kang2017patchshuffleregularization} and PatchMix~\cite{hong2024patchmix}, have influenced the design of our augmentation strategy for temporal data.

PatchMix divides an input image into non-overlapping patches, shuffles them, and then recombines the shuffled patches using a mixup strategy with weights sampled from a Beta distribution. An important feature of PatchMix is that the mixing process introduces patches that may bypass the mixing step and go directly to the final composition, making the resulting image more diverse~\cite{hong2024patchmix}.

\begin{figure}[ht]
  \begin{center}
    \centerline{\includegraphics[width=0.75\columnwidth]{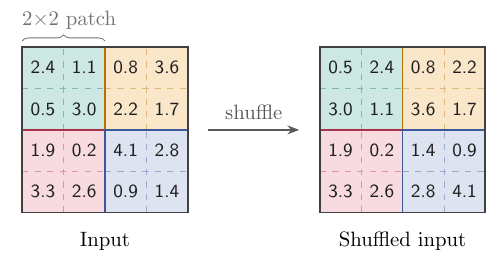}}
\caption{Illustration of PatchShuffle~\cite{kang2017patchshuffleregularization}.
A $4\times4$ input is divided into non-overlapping $2\times2$ patches,
with values independently shuffled within each patch while patch
positions remain unchanged.}
\Description{
Illustration of PatchShuffle on a four-by-four input divided into four
non-overlapping two-by-two patches. Values are independently permuted
within each patch while the positions of the four patches remain unchanged.
}
    \label{motiv-patchshuffle}
  \end{center}
\end{figure}

Figure~\ref{motiv-patchshuffle} presents a simplified version of the PatchShuffle method. In this example, a $4 \times 4$ image matrix is divided into four non-overlapping $2 \times 2$ patches. The pixels within each patch are independently shuffled: attributes in each patch are permuted separately, and a shuffled patch may also retain its original structure. This local pixel-level shuffling introduces variation while preserving the global structure of the image~\cite{kang2017patchshuffleregularization}.

Although PatchShuffle and PatchMix are effective in the CV domain, directly applying them to time series is not straightforward. The first challenge is that naive non-overlapping patching introduces boundary discontinuities that disrupt local temporal coherence, since time series are intrinsically sequential rather than arranged on a two-dimensional grid. The second challenge is forecasting-specific: augmentation should preserve coherence between the input window and its continuous future target, rather than perturbing the input alone. In addition, spatial transformations such as cropping, masking, or flipping, which are suitable for images, are generally inappropriate for sequential data unless carefully controlled.

To address these issues, TPS adapts the underlying intuition of patch-based augmentation to the temporal domain through a forecasting-tailored design. It uses overlapping patches and averaging-based reconstruction to make patch reordering less disruptive to local temporal structure, while variance-based ordering serves as a conservative heuristic when only a subset of patches is shuffled. In this way, TPS introduces controlled sample diversity while mitigating the temporal artifacts that naive CV-style patch shuffling would introduce.

\section{Approach}

\begin{figure*}[t]
\begin{center}
    \centerline{\includegraphics[width=\textwidth]{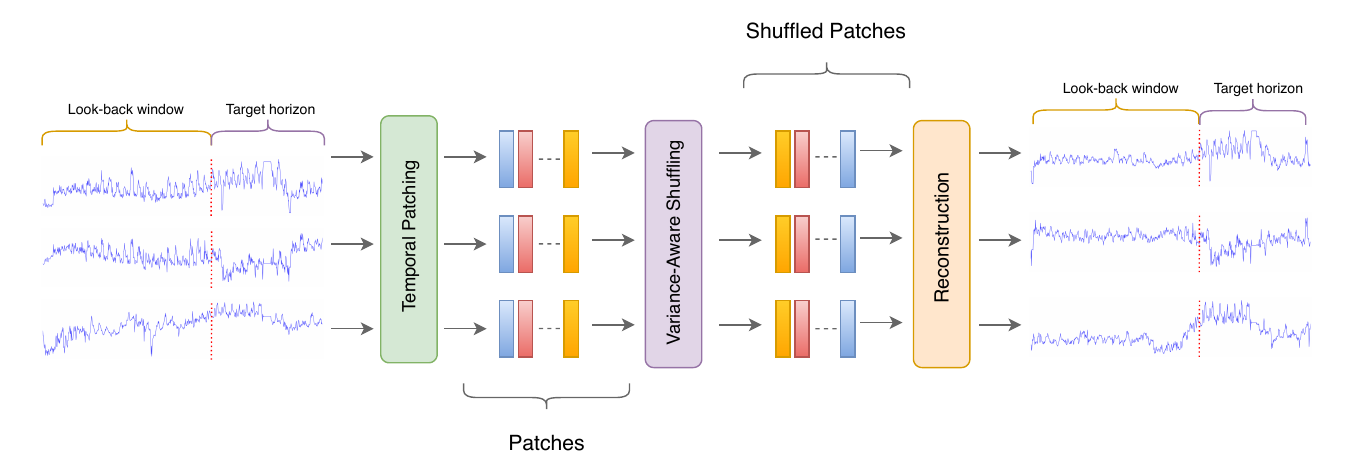}}
    \caption{Illustration of the proposed TPS method for time series forecasting. The input sequence, consisting of a look-back window and a forecast horizon, is first processed by the \textit{Temporal Patching} block to extract overlapping patches. These patches are then ordered by variance and partially shuffled in the \textit{Variance-Aware Shuffling} block. Finally, the shuffled patches are merged back into a full sequence via the \textit{Reconstruction} block by averaging overlapping regions.}
\Description{
Overview of Temporal Patch Shuffle. A time series containing a look-back
window and target horizon is divided into overlapping temporal patches.
Selected patches are reordered using variance-aware shuffling, and the
sequence is reconstructed by averaging overlapping regions.
}
    \label{fig:architecture}
\end{center}
\end{figure*}

Temporal Patch Shuffle (TPS) is a forecasting-tailored augmentation method that operates by extracting overlapping temporal patches, reordering a subset of them, and reconstructing the sequence by averaging overlapping regions. In the \textit{Temporal Patching} block (depicted in green in Figure~\ref{fig:architecture}), overlapping windows are extracted using the patch length and stride as hyperparameters. In the subsequent \textit{Variance-Aware Shuffling} block (depicted in purple in Figure~\ref{fig:architecture}), these patches are ordered by variance and selectively shuffled according to a predefined \textit{shuffle rate}, $\alpha$. Finally, the augmented sequence is reconstructed by averaging overlapping regions. Before applying the Temporal Patching block, the look-back window and forecast horizon are concatenated to preserve input--target alignment during augmentation.

\paragraph{Temporal Patching.}
Let a multivariate time series be defined as the concatenation of a look-back window $\mathbf{L}$ and a forecast horizon $\mathbf{F}$:
\begin{equation} \label{eq:concat2}
\mathbf{X} = [\mathbf{L}, \mathbf{F}] \in \mathbb{R}^{B \times T \times C}.
\end{equation}

The Temporal Patching block extracts overlapping patches of length $p$ with stride $s$:
\begin{equation} \label{eq:patching}
\mathbf{P} = \text{Unfold}(\mathbf{X}, p, s) \in \mathbb{R}^{B \times N_p \times C \times p},
\end{equation}
where the number of patches is
\begin{equation}
N_p = \left\lfloor \frac{T - p}{s} + 1 \right\rfloor.
\end{equation}

The \textit{Unfold} operation extracts patches (sliding windows) from the temporal dimension. The resulting patch tensor is denoted by $\mathbf{P} \in \mathbb{R}^{B \times N_p \times C \times p}$, where $B$ is the batch size, $N_p$ is the number of patches, $C$ is the number of channels, and $p$ is the patch length.

Each patch is formed by selecting a window of length $p$ along the temporal axis:
\begin{equation}
\mathbf{P}[b,i] = \Big(\mathbf{X}\big[b,\, i \cdot s : i \cdot s + p,\,: \big]\Big)^\top,
\end{equation}
where $b \in \{0,\dots,B-1\}$ and $i \in \{0,\dots,N_p-1\}$.

\paragraph{Variance-Aware Shuffling.}
To prioritize patches for shuffling when only a subset is reordered, TPS computes a variance-based score. This score is used as a simple conservative heuristic for perturbation ordering. Let $\mathbf{P}_b := \mathbf{P}[b, :, :, :] \in \mathbb{R}^{N_p \times C \times p}$ denote the patch set for batch element $b$. The mean value of each patch is
\begin{equation}
\bar{\mathbf{P}}[b, i] = \frac{1}{C \cdot p} \sum_{c=0}^{C-1} \sum_{j=0}^{p-1} \mathbf{P}[b, i, c, j],
\end{equation}
and the variance is
\begin{equation}
\text{Var}(\mathbf{P}[b,i])
= \frac{1}{C\cdot p - 1}\sum_{c,j}
\left( \mathbf{P}[b,i,c,j] - \bar{\mathbf{P}}[b,i] \right)^2 ,
\end{equation}
where $\sum_{c,j}$ is shorthand for $\sum_{c=0}^{C-1}\sum_{j=0}^{p-1}$ and the variance is well-defined for $C\cdot p>1$ (i.e., $p>1$ or $C>1$).

This yields the score vector:
\begin{equation} \label{eq:score}
\text{Score} = \text{Var}(\mathbf{P}) \in \mathbb{R}^{B \times N_p}.
\end{equation}

In our experiments, inputs are standardized per channel using statistics computed on the training split; variance scores are computed in this normalized space.

A fraction $\alpha$ of lowest-variance patches is selected for shuffling:
\begin{equation} \label{eq:numberofpatches}
N_s = \left\lfloor \alpha N_p \right\rfloor, \qquad \alpha \in (0,1].
\end{equation}
Let the selected patch indices for batch element $b$ be
\begin{equation}\label{eq:patch_indices}
\mathcal{I}_b \subseteq \{0,\dots,N_p-1\}, \qquad |\mathcal{I}_b| = N_s,
\end{equation}
where $\mathcal{I}_b$ contains the indices of the $N_s$ smallest values in $\text{Score}[b,:]$.

The chosen patches are then randomly permuted within each batch element:
\begin{equation}\label{eq:patchshuffling}
\mathbf{P}_b[\mathcal{I}_b] \leftarrow \mathbf{P}_b[\mathcal{I}_b][\pi_b],
\end{equation}
where $\pi_b$ is a random permutation of $\{0,\dots,N_s-1\}$ that reorders the $N_s$ selected patches.

\begin{algorithm}[tb]
  \caption{Temporal Patch Shuffle (TPS)}
  \label{alg:tps}
  \begin{algorithmic}[1]
    \REQUIRE look-back window $\mathbf{L}$, forecast horizon $\mathbf{F}$, patch length $p$, stride $s$, shuffle rate $\alpha$
    \ENSURE augmented sequence $\mathbf{S}$
    \STATE $\mathbf{X} \leftarrow [\mathbf{L}, \mathbf{F}]$ \COMMENT{$\mathbf{X}\in\mathbb{R}^{B\times T\times C}$}
    \STATE $\mathbf{P} \leftarrow \text{Unfold}(\mathbf{X}, p, s)$ \COMMENT{$\mathbf{P}\in\mathbb{R}^{B\times N_p\times C\times p}$}
    \STATE $\text{Score} \leftarrow \text{Var}(\mathbf{P})$ \COMMENT{$\text{Score}\in\mathbb{R}^{B\times N_p}$}
    \STATE $N_s \leftarrow \lfloor \alpha N_p \rfloor$
    \FOR{$b = 0$ to $B-1$}
      \STATE $\mathcal{I}_b \leftarrow \text{Argsort}(\text{Score}[b,:])[0:N_s]$
      \STATE $\pi_b \leftarrow \text{RandPerm}(N_s)$
      \STATE $\mathbf{P}_b[\mathcal{I}_b] \leftarrow \mathbf{P}_b[\mathcal{I}_b][\pi_b]$
    \ENDFOR
    \STATE $\widetilde{\mathbf{P}} \leftarrow \mathbf{P}$ \COMMENT{shuffled patch tensor}
    \STATE $\mathbf{S} \leftarrow \text{Reconstruct}(\widetilde{\mathbf{P}})$
  \end{algorithmic}
\end{algorithm}

\paragraph{Reconstruction.}
The Reconstruction block places each (shuffled) patch back at its original temporal position and averages overlapping regions. Let $\widetilde{\mathbf{P}}$ denote the patch tensor after shuffling. The reconstructed sequence $\mathbf{S}\in\mathbb{R}^{B\times T\times C}$ is defined element-wise as
\begin{equation}
\mathbf{S}[b,\tau,c]
=
\frac{1}{K_\tau}
\sum_{i=0}^{N_p-1}
\mathbb{I}\{\tau \in [i \cdot s,\, i \cdot s + p)\}\;
\widetilde{\mathbf{P}}[b,i,c,\tau-i \cdot s],
\end{equation}
for $b\in\{0,\dots,B-1\}$, $\tau\in\{0,\dots,T-1\}$, and $c\in\{0,\dots,C-1\}$,
where
\begin{equation}
K_\tau = \sum_{i=0}^{N_p-1}\mathbb{I}\{\tau \in [i \cdot s,\, i \cdot s + p)\}.
\end{equation}
This explicitly accounts for boundary regions, which are covered by fewer overlapping patches.

Finally, $\mathbf{S}$ is partitioned into the augmented look-back window and forecast horizon:
\begin{equation}
\begin{aligned}
\mathbf{S}_L &= \mathbf{S}[:,\,0:t,\,:] \in \mathbb{R}^{B \times t \times C},\\
\mathbf{S}_F &= \mathbf{S}[:,\,t:T,\,:] \in \mathbb{R}^{B \times h \times C}.
\end{aligned}
\end{equation}

They are concatenated with their corresponding original samples, as in Eq.~(\ref{eq:concat}). These follow the same training pipeline illustrated in Figure~\ref{fig:tsf-frame}. The complete TPS procedure is outlined in Algorithm~\ref{alg:tps}.

\section{Experiments}

We first describe our experimental settings (Section~\ref{subsection:experiment_settings}) and then evaluate TPS on widely used benchmark datasets for long-term forecasting (Section~\ref{subsection:longterm}) and short-term forecasting (Section~\ref{subsection:shortterm}). We also provide in-depth analyses examining component-wise contributions, distributional behavior, hyperparameter sensitivity, probabilistic forecasting quality, computational overhead, and extensions beyond forecasting (Section~\ref{subsection:ablation}).

\subsection{Experimental Settings} \label{subsection:experiment_settings}

We summarize the key experimental settings below; additional details are provided in the supplementary material.

\paragraph{Models and Datasets.}
We evaluate TPS across a diverse set of forecasting architectures, ranging from linear models and MLP-based designs to transformer-style models, including DLinear~\cite{zeng2022transformerseffectivetimeseries}, TSMixer~\cite{chen2023tsmixerallmlparchitecturetime}, TiDE~\cite{das2024longtermforecastingtidetimeseries}, LightTS~\cite{zhang2022morefastmultivariatetime}, and the transformer-based PatchTST~\cite{nie2023timeseriesworth64}. All models are trained using their respective default hyperparameters from the original implementations.

For evaluation, we use nine long-term and four short-term forecasting datasets, summarized in Table~\ref{tab:datasets}. The ETT datasets~\cite{zhou2021informerefficienttransformerlong} comprise ETTh1, ETTh2, ETTm1, and ETTm2, while Exchange, Weather, ECL, and Traffic~\cite{wu2022autoformerdecompositiontransformersautocorrelation} cover economic, meteorological, electricity-consumption, and traffic domains, respectively. We also include the weekly Influenza-like Illness (ILI) dataset. For short-term forecasting, we use PeMS03, PeMS04, PeMS07, and PeMS08, following the SCINet protocol~\cite{liu2022scinettimeseriesmodeling}. Together, these benchmarks span diverse domains, sampling frequencies, and dimensionalities~\cite{liu2024itransformerinvertedtransformerseffective}.

\begin{table}[ht]
\centering
\footnotesize
\renewcommand{\arraystretch}{1.1}
\setlength{\tabcolsep}{3pt}
\caption{Summary of datasets used in experiments, including dimensionality, prediction lengths, and dataset sizes (train, val, test). The top block corresponds to long-term forecasting and the bottom block to short-term forecasting. Sampling frequencies are abbreviated in parentheses (H: hourly, D: daily, W: weekly, 10m/15m/5m: minutes).}
\label{tab:datasets}
\begin{tabular}{lcccc}
\toprule
\textbf{Dataset} & \textbf{Dim.} & \textbf{Pred.\ Length} & \textbf{Size} & \textbf{Info} \\
\midrule
ETTh1 & 7 & \{96,192,336,720\} & (8545,2881,2881) & Elec.\ (H) \\
ETTh2 & 7 & \{96,192,336,720\} & (8545,2881,2881) & Elec.\ (H) \\
ETTm1 & 7 & \{96,192,336,720\} & (34465,11521,11521) & Elec.\ (15m) \\
ETTm2 & 7 & \{96,192,336,720\} & (34465,11521,11521) & Elec.\ (15m) \\
Exchange & 8 & \{96,192,336,720\} & (5120,665,1422) & Econ.\ (D) \\
Weather & 21 & \{96,192,336,720\} & (36792,5271,10540) & Weather (10m) \\
ECL & 321 & \{96,192,336,720\} & (18317,2633,5261) & Elec.\ (H) \\
Traffic & 862 & \{96,192,336,720\} & (12185,1757,3509) & Transp.\ (H) \\
ILI & 7 & \{24,36,48,60\} & (629,98,194) & Illness (W) \\
\midrule
PeMS03 & 358 & \{12,24,36,48\} & (15617,5135,5135) & Transp.\ (5m) \\
PeMS04 & 307 & \{12,24,36,48\} & (10172,3375,3375) & Transp.\ (5m) \\
PeMS07 & 883 & \{12,24,36,48\} & (16911,5622,5622) & Transp.\ (5m) \\
PeMS08 & 170 & \{12,24,36,48\} & (10690,3548,3548) & Transp.\ (5m) \\
\bottomrule
\end{tabular}
\end{table}

\begin{table*}[t]
\caption{
Long-term forecasting performance (MSE/MAE) averaged over nine datasets and four prediction lengths.
\#Wins is counted over 36 settings (9 datasets $\times$ 4 prediction lengths).
Best results are highlighted in bold, and second-best results are underlined.
STAug results are averaged over seven datasets due to GPU memory limitations. Matched seven-dataset averages (excluding ECL and Traffic) and per-dataset standard deviations are reported in the supplementary material.
}
\label{tab:mainresult1}
\centering
\footnotesize
\setlength{\tabcolsep}{5pt}
\renewcommand{\arraystretch}{1.2}
\begin{tabular}{c|cc|cc|cc|cc|cc}
\toprule
\textbf{Method} &
\multicolumn{2}{c|}{\textbf{TSMixer}} &
\multicolumn{2}{c|}{\textbf{DLinear}} &
\multicolumn{2}{c|}{\textbf{PatchTST}} &
\multicolumn{2}{c|}{\textbf{TiDE}} &
\multicolumn{2}{c}{\textbf{LightTS}} \\
& MSE & MAE & MSE & MAE & MSE & MAE & MSE & MAE & MSE & MAE \\
\midrule
None & 0.461 & 0.403 & 0.548 & 0.439 & 0.468 & 0.399 & 0.483 & 0.409 & 0.649 & 0.482 \\
wDBA~\cite{asd}  & 0.462 & 0.403 & 0.536 & 0.433 & 0.459 & 0.396 & 0.495 & 0.417 & 0.638 & 0.477 \\
MBB~\cite{mbb}   & 0.467 & 0.405 & 0.544 & 0.439 & 0.459 & \secondres{0.395} & 0.495 & 0.416 & 0.651 & 0.484 \\
RobustTAD-m/p~\cite{gao2021robusttadrobusttimeseries} & 0.462 & \secondres{0.401} & 0.541 & 0.436 & 0.464 & 0.404 & 0.485 & 0.409 & 0.651 & 0.481 \\
FreqAdd~\cite{freqadd} & \secondres{0.459} & 0.402 & 0.556 & 0.443 & 0.467 & 0.400 & 0.481 & 0.408 & 0.645 & 0.481 \\
FreqPool~\cite{freqpool} & 0.473 & 0.408 & \secondres{0.533} & 0.437 & 0.476 & 0.403 & 0.499 & 0.419 & 0.624 & 0.471 \\
Upsample~\cite{upsample}  & 0.477 & 0.410 & 0.535 & 0.440 & 0.468 & 0.399 & 0.516 & 0.423 & \secondres{0.609} & \secondres{0.464} \\
STAug~\cite{zhang2023diversecoherentaugmentationtimeseries}   & 0.540 & 0.451 & 0.733 & 0.538 & 0.548 & 0.453 & 0.630 & 0.497 & 0.855 & 0.583 \\
Freq-Mask/Mix~\cite{chen2023fraugfrequencydomainaugmentation} & 0.468 & 0.405 & 0.540 & \secondres{0.432} & 0.462 & 0.400 & 0.492 & 0.412 & 0.636 & 0.475 \\
Wave-Mask/Mix~\cite{arabi2024wavemaskmixexploringwaveletbasedaugmentations} & 0.463 & 0.403 & 0.545 & 0.436 & \secondres{0.458} & 0.398 & \secondres{0.480} & \secondres{0.407} & 0.643 & 0.478 \\
Dominant Shuffle~\cite{zhao2024dominantshufflesimplepowerful} & 0.471 & 0.407 & 0.545 & 0.433 & 0.469 & 0.402 & 0.493 & 0.411 & 0.634 & 0.472 \\
\rowcolor{tpsblue} \textbf{TPS (Ours)}   & \boldres{0.447} & \boldres{0.394} & \boldres{0.493} & \boldres{0.410} &
\boldres{0.445} & \boldres{0.388} & \boldres{0.470} & \boldres{0.401} &
\boldres{0.545} & \boldres{0.438} \\
\midrule
\textbf{\#Wins (/36)} & 26 & 29 & 35 & 34 & 27 & 27 & 30 & 27 & 32 & 32 \\
\rowcolor{tpsblue}
\textbf{Improvement}
& \boldres{2.61\%} & \boldres{1.75\%} & \boldres{7.50\%} & \boldres{5.09\%} & \boldres{2.84\%} & \boldres{1.77\%} & \boldres{2.08\%} & \boldres{1.47\%} & \boldres{10.51\%} & \boldres{5.60\%} \\
\bottomrule
\end{tabular}
\end{table*}

\begin{table*}[t]
\caption{
Short-term traffic forecasting with PatchTST on PeMS-\{03, 04, 07, 08\}.
Mean MSE/MAE are computed over 5 runs per prediction length and then averaged over $\{12,24,36,48\}$.
Best results are highlighted in bold, and second-best results are underlined. Standard deviations are reported in the supplementary material.
}
\label{tab:main_result2}
\centering
\footnotesize
\renewcommand{\arraystretch}{1.2}
\setlength{\tabcolsep}{5pt}
\begin{tabular}{c|cc|cc|cc|cc}
\toprule
\multicolumn{1}{c|}{\multirow{2}{*}{\textbf{Method}}} &
\multicolumn{2}{c|}{\textbf{PeMS03}} &
\multicolumn{2}{c|}{\textbf{PeMS04}} &
\multicolumn{2}{c|}{\textbf{PeMS07}} &
\multicolumn{2}{c}{\textbf{PeMS08}} \\
& MSE & MAE & MSE & MAE & MSE & MAE & MSE & MAE \\
\midrule
None
& 0.118 & 0.234
& 0.135 & 0.246
& 0.117 & 0.241
& 0.159 & 0.259\\
wDBA~\cite{asd}
& 0.125 & 0.247
& 0.134 & 0.246
& 0.109 & 0.233
& 0.170 & 0.262 \\
MBB~\cite{mbb}
& 0.118 & 0.231
& 0.139 & 0.255
& 0.125 & 0.258
& 0.169 & 0.260 \\
RobustTAD-m/p~\cite{gao2021robusttadrobusttimeseries}
& 0.122 & 0.240
& 0.134 & 0.249
& 0.111 & 0.235
& 0.159 & 0.259 \\
FreqAdd~\cite{freqadd}
& 0.123 & 0.241
& 0.139 & 0.252
& 0.118 & 0.249
& 0.161 & 0.268 \\
FreqPool~\cite{freqpool}
& \secondres{0.112} & 0.230
& \secondres{0.128} & \secondres{0.243}
& 0.108 & 0.232
& 0.143 & 0.252 \\
Upsample~\cite{upsample}
& 0.112 & 0.231
& 0.136 & 0.254
& 0.109 & 0.233
& \secondres{0.141} & \secondres{0.248} \\
STAug~\cite{zhang2023diversecoherentaugmentationtimeseries}
& 0.113 & \secondres{0.224}
& 0.135 & 0.249
& 0.118 & 0.235
& 0.150 & 0.251\\
Freq-Mask/Mix~\cite{chen2023fraugfrequencydomainaugmentation}
& 0.124 & 0.243
& 0.143 & 0.257
& 0.113 & 0.235
& 0.167 & 0.266 \\
Wave-Mask/Mix~\cite{arabi2024wavemaskmixexploringwaveletbasedaugmentations}
& 0.115 & 0.228
& 0.133 & 0.246
& \boldres{0.105} & \secondres{0.228}
& 0.149 & 0.258 \\
Dominant Shuffle~\cite{zhao2024dominantshufflesimplepowerful}
& 0.115 & 0.231
& 0.134 & 0.252
& 0.109 & 0.231
& 0.156 & 0.256\\
\rowcolor{tpsblue} \textbf{TPS (Ours)}
& \boldres{0.104} & \boldres{0.216}
& \boldres{0.125} & \boldres{0.238}
& \boldres{0.105} & \boldres{0.225}
& \boldres{0.135} & \boldres{0.240} \\
\midrule
\textbf{\#Wins (/4)} & 3 & 4 &  3 & 3 &   3 & 3 & 3 & 3   \\

\rowcolor{tpsblue} \textbf{Improvement}
& \boldres{7.14\%} & \boldres{3.57\%}
& \boldres{2.34\%} & \boldres{2.06\%}
& 0.00\% & \boldres{1.32\%}
& \boldres{4.26\%} & \boldres{3.23\%} \\
\bottomrule
\end{tabular}
\end{table*}

\paragraph{Baselines.}
We compare TPS against a comprehensive set of existing augmentation methods, including wDBA~\cite{asd}, MBB~\cite{mbb}, RobustTAD~\cite{gao2021robusttadrobusttimeseries}, FreqAdd~\cite{freqadd}, FreqPool~\cite{freqpool}, Upsample~\cite{upsample}, STAug~\cite{zhang2023diversecoherentaugmentationtimeseries}, FreqMask/FreqMix~\cite{chen2023fraugfrequencydomainaugmentation}, WaveMask/WaveMix~\cite{arabi2024wavemaskmixexploringwaveletbasedaugmentations}, and Dominant Shuffle~\cite{zhao2024dominantshufflesimplepowerful}. For STAug, FreqMask/FreqMix, WaveMask/WaveMix, and Dominant Shuffle, we use the official implementations released by the respective authors; the remaining baselines are reproduced following their published descriptions and recommended configurations.

Augmentations based on generative models (e.g., TimeGAN~\cite{timegan}) are omitted because they require training additional generator models and are substantially more computationally expensive; moreover, prior work reports limited benefits and potential noise artifacts~\cite{chen2023fraugfrequencydomainaugmentation, zhang2023diversecoherentaugmentationtimeseries}.

\begin{figure*}[ht]
\begin{center}
    \centerline{\includegraphics[width=0.95\textwidth]{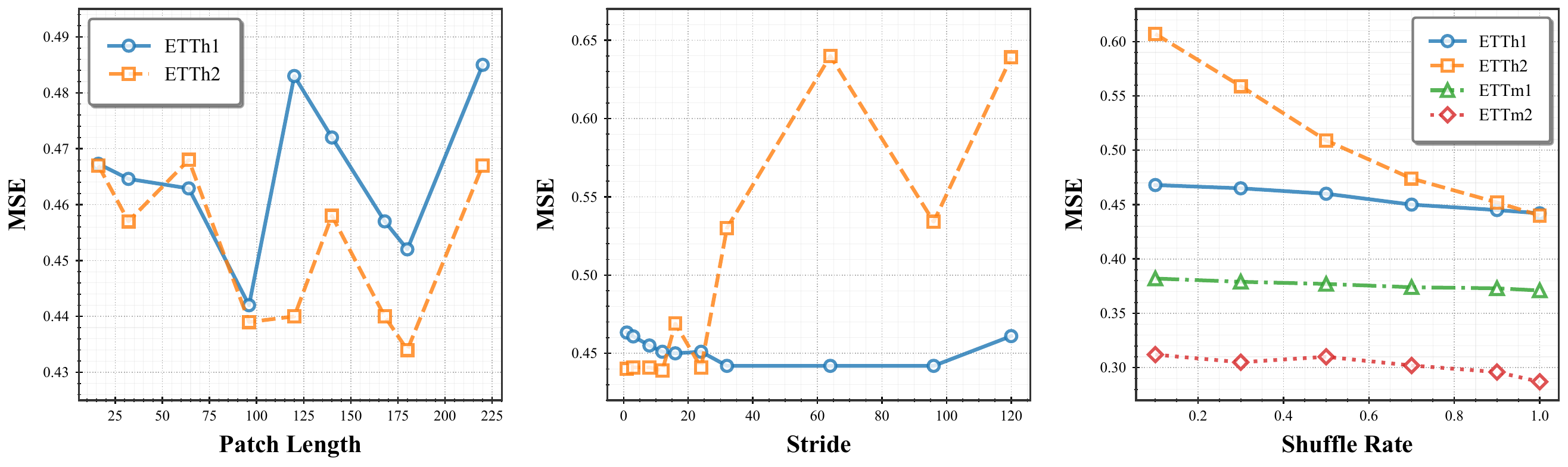}}
\caption{
Hyperparameter sensitivity of TPS on ETT with LightTS (prediction length 336). MSE averaged over five runs while varying one of $\{p,s,\alpha\}$ and fixing the other two to the best validation configuration.
}
\Description{
Three line plots showing the sensitivity of TPS forecasting error to
patch length, stride, and shuffle rate on the ETT datasets. Performance
varies non-monotonically with patch length and stride, while higher
shuffle rates generally reduce MSE.
}
	\label{fig:hyper_sensitivity}
\end{center}
\end{figure*}

\paragraph{Experimental Setups.}
Input sequence lengths follow the recommended configurations from the original backbone papers. All models are trained for 20 epochs (or fewer if the original configuration specifies a smaller value), with early stopping (patience 10) and the checkpoint selected by the lowest validation loss. Learning-rate schedulers are chosen according to the recommendations of the original papers when available, or otherwise based on empirical validation to ensure stable convergence. Learning rates are set using the original configurations as a starting point and adjusted when necessary based on validation performance. We select TPS hyperparameters $(p,s,\alpha)$ via validation-based search over approximately 20 candidate combinations. This unified protocol allows us to isolate augmentation effects under identical optimization conditions across all augmentation methods and backbone models. As a result, our reproduced numbers may differ slightly from those reported in the original papers. For instance, we adopt a shorter training schedule for PatchTST and observe performance broadly comparable to the original setup, with only minor deviations at some prediction lengths; we further verified that our results remain stable under longer training schedules.

\subsection{Main Results}

\subsubsection{Long-Term Forecasting}\label{subsection:longterm}

Table~\ref{tab:mainresult1} reports average performance on nine datasets and four prediction lengths using MSE and MAE. For each dataset and prediction length, results are averaged over 5 runs; we then average across the four prediction lengths and across datasets, yielding 36 settings (9 datasets~$\times$~4 prediction lengths). RobustTAD-m/p denotes the best result among RobustTAD variants; Freq-Mask/Mix and Wave-Mask/Mix denote the best outcomes among the corresponding variants. STAug results are averaged over seven datasets due to GPU memory limitations (consistent with the original STAug paper~\cite{zhang2023diversecoherentaugmentationtimeseries}).

TPS achieves substantial improvements of \textbf{2.61\%}, \textbf{7.50\%}, \textbf{2.84\%}, \textbf{2.08\%}, and \textbf{10.51\%} in MSE for TSMixer, DLinear, PatchTST, TiDE, and LightTS, respectively, while also obtaining a high number of wins across all models. Per-dataset results with standard deviations are provided in the supplementary material.

STAug could not be evaluated on ECL and Traffic due to GPU memory limitations, consistent with the original STAug paper~\cite{zhang2023diversecoherentaugmentationtimeseries}. Accordingly, its averages in Table~\ref{tab:mainresult1} are computed over the seven datasets on which it is runnable. For a strictly matched comparison across all methods, seven-dataset averages are reported in the supplementary material.

\subsubsection{Short-Term Forecasting}\label{subsection:shortterm}

Table~\ref{tab:main_result2} presents results on the PeMS datasets using PatchTST. For each dataset and prediction length, metrics are averaged over 5 runs, and we report the average over prediction lengths $\{12,24,36,48\}$. TPS obtains the best or tied-best average MSE on all four datasets and the best average MAE on all four. Specifically, it improves MSE by \textbf{7.14\%}, \textbf{2.34\%}, and \textbf{4.26\%} on PeMS-\{03, 04, 08\}, respectively, while matching the best MSE on PeMS07 and improving its MAE by \textbf{1.32\%}. Per-prediction-length results with standard deviations are provided in the supplementary material.

\subsection{Further Analysis and Extensions} \label{subsection:ablation}

We conducted several ablation studies to analyze the design choices of TPS. Additional details and extended results, including augmentation size and augmentation ratio sensitivity analyses, are provided in the supplementary material.

\paragraph{Component-wise Analysis.}
Table~\ref{tab:ablation_dlinear} reports a component-wise ablation of TPS with DLinear~\cite{zeng2022transformerseffectivetimeseries} on the ETT datasets, averaged over four prediction lengths $\{96,192,336,720\}$. We report MSE here; the corresponding MAE results and standard deviations are provided in the supplementary material. Removing variance-based sorting produces a modest degradation overall, indicating that variance-aware selection acts as a refinement when only a subset of patches is shuffled rather than as the primary source of TPS's gains. Its effect naturally disappears for configurations with $\alpha=1$, where all patches are selected. Replacing overlapping patches with non-overlapping ones causes a clearer degradation, supporting overlap-averaged reconstruction as an important mechanism for limiting temporal discontinuities. Applying augmentation only to the input (breaking data--label coherence) substantially hurts performance, consistent with prior observations~\cite{chen2023fraugfrequencydomainaugmentation, wei2020circumventingoutliersautoaugmentknowledge}. A frequency-domain variant, where the same patch-based operations are applied after the Fast Fourier Transform, also degrades results, suggesting that this construction is more effective directly in the time domain.

\begin{table}[ht]
\caption{
Component-wise ablation (MSE) of TPS on ETT with DLinear, averaged over four prediction lengths. 
Best is highlighted in bold, and second-best is underlined. MAE results and standard deviations are provided in the supplementary material.
}
\label{tab:ablation_dlinear}
\centering
\footnotesize
\renewcommand{\arraystretch}{1.0}
\setlength{\tabcolsep}{3pt}
\begin{tabular}{l|cccc}
\toprule
\textbf{Method} & \textbf{ETTh1} & \textbf{ETTh2} & \textbf{ETTm1} & \textbf{ETTm2} \\
\midrule
None & 0.438 & 0.464 & 0.361 & 0.276 \\
\rowcolor{tpsblue} \textbf{TPS} & \boldres{0.410} & \boldres{0.369} & \boldres{0.354} & \boldres{0.261} \\
\hspace{0.5em}- Variance Score
& 0.417 & \secondres{0.370} & \secondres{0.355} & \boldres{0.261} \\
\hspace{0.5em}- Temporal Patching
& \secondres{0.416} & 0.379 & 0.376 & \secondres{0.267} \\
\hspace{0.5em}- Data-Label Coherence
& 0.443 & 0.438 & 0.364 & 0.290 \\
\hspace{0.5em}+ Frequency Dom.
& 0.437 & 0.470 & 0.363 & 0.285 \\
\bottomrule
\end{tabular}
\end{table}

\paragraph{Hyperparameter Sensitivity.}
Figure~\ref{fig:hyper_sensitivity} presents the sensitivity of TPS to its hyperparameters. Using LightTS with a prediction length of 336, we varied patch length, stride, and shuffle rate on the ETT datasets, fixing the other two to the best validation configuration. Higher shuffle rates consistently reduced MSE, so values in the range of 0.7--1.0 were typically chosen. Stride and patch length exhibit non-monotonic, dataset-dependent trends. Overall, smaller strides are typically more stable, and moderately larger patch lengths often outperform very small ones. When $\alpha=1.0$, all patches are shuffled and variance ordering becomes irrelevant by construction; for validation-selected configurations with $\alpha<1$, the variance criterion instead determines which subset is reordered. Thus, the strong performance observed at high shuffle rates primarily reflects the overlapping-patch reordering and reconstruction pipeline, with variance-aware selection providing an additional refinement under partial shuffling.

\begin{figure*}[ht]
\begin{center}
    \centerline{\includegraphics[width=\textwidth]{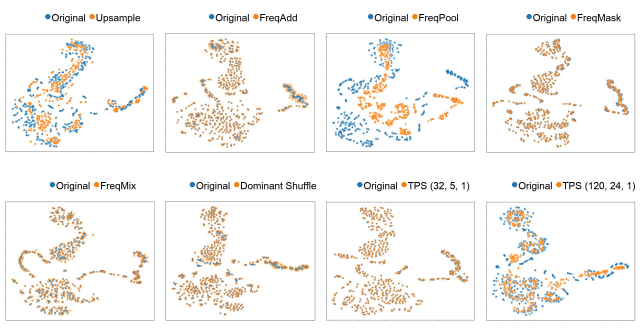}}
\caption{t-SNE visualization of original and augmented data on ETTh2 using DLinear with prediction length 336. Closer overlap between original and augmented points suggests better distributional alignment. \textbf{Best viewed in color.}}
\Description{
Eight t-SNE plots comparing original ETTh2 samples with samples produced
by several augmentation methods. The validation-selected TPS configuration
shows strong overlap with the original samples, while an alternative TPS
configuration produces greater variation while retaining the broad structure.
}
    \label{fig:tsne}
\end{center}
\end{figure*}

\paragraph{Distributional-Alignment Analysis.}
Table~\ref{tab:dist_metrics} reports a quantitative distributional-alignment comparison between original and augmented samples on ETTh2 using DLinear with a prediction length of 336. We compute three metrics: the Kolmogorov--Smirnov (KS) statistic, Wasserstein distance, and Dynamic Time Warping (DTW) distance~\cite{gchron-5-263-2023, Iglesias2023}. For KS and Wasserstein, we compute the metric per channel by flattening across batch and time, then average across channels. For DTW, we compute sequence-level distances per sample and channel and average over all.

TPS with the validation-selected configuration $(p,s,\alpha)=(32,5,1.0)$ achieves the lowest Wasserstein distance (0.0097) and DTW distance (1.46), indicating strong preservation of both distributional structure and temporal dynamics. While TPS has a slightly higher KS statistic than Dominant Shuffle, the substantially lower DTW distance (1.46 vs.\ 4.91--14.72 for competitors) supports that TPS mitigates temporal distortion more effectively. Figure~\ref{fig:tsne} presents a t-SNE visualization confirming these trends: the validation-selected configuration $(32,5,1)$ produces augmented samples that overlap most closely with the original data, while an alternative configuration $(120,24,1)$ introduces stronger variation while still preserving broad structural coherence.

\begin{table}[ht]
\caption{
Distributional-alignment comparison between original and augmented samples on ETTh2 (DLinear, prediction length 336).
Lower values indicate closer alignment between original and augmented data. Best in bold.
}

\label{tab:dist_metrics}
\centering
\footnotesize
\setlength{\tabcolsep}{3pt}
\renewcommand{\arraystretch}{1.1}
\begin{tabular}{lccc}
\toprule
\textbf{Method}
& \textbf{KS Stat} $\downarrow$
& \textbf{Wasserstein} $\downarrow$
& \textbf{DTW} $\downarrow$ \\
\midrule
Upsample~\cite{upsample}         & 0.0202 & 0.0177 & 8.73 \\
FreqAdd~\cite{freqadd}          & 0.1019 & 0.1475 & 8.55 \\
FreqPool~\cite{freqpool}         & 0.3366 & 0.3839 & 14.72 \\
FreqMask~\cite{chen2023fraugfrequencydomainaugmentation} & 0.0793 & 0.0523 & 4.91 \\
FreqMix~\cite{chen2023fraugfrequencydomainaugmentation}  & 0.0756 & 0.0855 & 7.12 \\
Dominant Shuffle~\cite{zhao2024dominantshufflesimplepowerful} & \boldres{0.0688} & 0.0550 & 6.02 \\
\textbf{TPS (32,5,1)} & 0.0848 & \boldres{0.0097} & \boldres{1.46} \\
\bottomrule
\end{tabular}
\end{table}

\paragraph{Probabilistic Forecasting.}
We evaluate TPS in a probabilistic forecasting setting using quantile regression with nine quantiles ($\tau \in \{0.05, 0.1, 0.2, 0.3, 0.5, 0.7, 0.8, 0.9, 0.95\}$) and DLinear on the four ETT datasets across all four prediction lengths.
Table~\ref{tab:prob_forecasting} reports pinball loss, the Continuous Ranked Probability Score (CRPS)~\cite{gneiting2007strictly}, and 80\% prediction-interval (PI-80\%) coverage and width, averaged over prediction lengths $\{96, 192, 336, 720\}$ and 5 seeds.
TPS improves pinball loss and CRPS on all four datasets when averaged, while producing substantially sharper prediction intervals. PI-80\% coverage is more mixed, with TPS tending to slightly under-cover and the no-augmentation baseline tending to slightly over-cover, consistent with the absence of post-hoc calibration.
This reflects a calibration--sharpness tradeoff: TPS produces sharper intervals and improves proper scoring metrics, while some settings may benefit from post-hoc calibration. Overall, these results show that TPS remains effective beyond the standard point-forecasting setup.

\begin{table}[t]
\caption{Probabilistic forecasting results (DLinear) on ETT datasets, averaged over prediction lengths $\{96, 192, 336, 720\}$ and 5 seeds.
Metrics: Pinball loss~($\downarrow$), CRPS~($\downarrow$), PI-80\% Coverage (nominal 0.80), and PI-80\% Width~($\downarrow$).
Best is highlighted in bold, and second-best is underlined. Coverage is reported without highlighting since values closer to nominal (not lower) are preferred.}
\label{tab:prob_forecasting}
\centering
\footnotesize
\setlength{\tabcolsep}{2pt}
\renewcommand{\arraystretch}{1.1}
\begin{tabular}{l|cccc|cccc}
\toprule
 & \multicolumn{4}{c|}{\textbf{ETTh1}} & \multicolumn{4}{c}{\textbf{ETTh2}} \\
\textbf{Method} & Pin. & CRPS & Cov. & Wid. & Pin. & CRPS & Cov. & Wid. \\
\midrule
None          & .1752 & .3504 & .828 & \secondres{1.389} & .1764 & .3527 & .791 & 1.356 \\
Upsample~\cite{upsample}      & .1778 & .3555 & .823 & 1.415 & .1693 & .3387 & .757 & \secondres{1.159} \\
FreqMask~\cite{chen2023fraugfrequencydomainaugmentation}      & .1788 & .3575 & .866 & 1.642 & .1718 & .3436 & .782 & 1.285 \\
FreqAdd~\cite{freqadd}       & .1785 & .3570 & .874 & 1.681 & .1827 & .3654 & .852 & 1.710 \\
FreqMix~\cite{chen2023fraugfrequencydomainaugmentation}       & .1774 & .3546 & .842 & 1.473 & .1743 & .3486 & .822 & 1.494 \\
Dominant Shuffle~\cite{zhao2024dominantshufflesimplepowerful} & \secondres{.1753} & \secondres{.3505} & .842 & 1.458 & \secondres{.1723} & \secondres{.3446} & .800 & 1.373 \\
\rowcolor{tpsblue} \textbf{TPS (Ours)}    & \boldres{.1727} & \boldres{.3454} & .792 & \boldres{1.232} & \boldres{.1636} & \boldres{.3273} & .739 & \boldres{1.024} \\
\midrule
 & \multicolumn{4}{c|}{\textbf{ETTm1}} & \multicolumn{4}{c}{\textbf{ETTm2}} \\
\textbf{Method} & Pin. & CRPS & Cov. & Wid. & Pin. & CRPS & Cov. & Wid. \\
\midrule
None          & .1556 & .3112 & .824 & 1.171 & .1341 & .2681 & .816 & 1.074 \\
Upsample~\cite{upsample}      & .1648 & .3296 & .802 & 1.183 & .1345 & .2690 & .772 & \secondres{0.918} \\
FreqMask~\cite{chen2023fraugfrequencydomainaugmentation}      & .1560 & .3119 & .812 & \secondres{1.132} & \secondres{.1340} & \secondres{.2680} & .804 & 1.016 \\
FreqAdd~\cite{freqadd}       & .1589 & .3178 & .864 & 1.393 & .1406 & .2812 & .884 & 1.478 \\
FreqMix~\cite{chen2023fraugfrequencydomainaugmentation}       & .1572 & .3144 & .842 & 1.266 & .1352 & .2705 & .826 & 1.129 \\
Dominant Shuffle~\cite{zhao2024dominantshufflesimplepowerful} & \secondres{.1570} & \secondres{.3140} & .836 & 1.238 & .1346 & .2693 & .829 & 1.133 \\
\rowcolor{tpsblue} \textbf{TPS (Ours)}    & \boldres{.1547} & \boldres{.3094} & .786 & \boldres{1.033} & \boldres{.1333} & \boldres{.2666} & .752 & \boldres{0.835} \\
\bottomrule
\end{tabular}
\end{table}

\paragraph{Computational Overhead.}
Table~\ref{tab:augmentation_comparison} reports training-time overhead using TSMixer on ETTh2 at prediction length 720. TPS incurs 78\% overhead over the no-augmentation baseline, substantially less than Dominant Shuffle at 235\%.

\begin{table}[ht]
\centering
\footnotesize
\renewcommand{\arraystretch}{1.1}
\caption{
Training-time comparison (TSMixer, ETTh2, pred.\ len.\ 720).
\textit{Overhead} is the \% increase in epoch time vs.\ None.
}
\label{tab:augmentation_comparison}
\setlength{\tabcolsep}{4pt}
\begin{tabular}{lccc}
    \toprule
    \textbf{Method} & \textbf{Aug.\ (ms)} & \textbf{Epoch (s)} & \textbf{Overhead} \\
    \midrule
    None  & 0.0 & 2.30 & 0\% \\
    \midrule
    FreqPool~\cite{freqpool}       & 1.0 & 2.61 & 14\% \\
    FreqMask~\cite{chen2023fraugfrequencydomainaugmentation}       & 1.1 & 2.63 & 15\% \\
    FreqAdd~\cite{freqadd}        & 1.2 & 2.64 & 15\% \\
    FreqMix~\cite{chen2023fraugfrequencydomainaugmentation}       & 1.6 & 2.72 & 18\% \\
    RobustTAD-m~\cite{gao2021robusttadrobusttimeseries} & 1.6 & 2.69 & 17\% \\
    RobustTAD-p~\cite{gao2021robusttadrobusttimeseries} & 1.7 & 2.76 & 20\% \\
    WaveMix~\cite{arabi2024wavemaskmixexploringwaveletbasedaugmentations}        & 2.3 & 2.86 & 25\% \\
    Upsample~\cite{upsample}       & 2.5 & 2.94 & 28\% \\
    WaveMask~\cite{arabi2024wavemaskmixexploringwaveletbasedaugmentations}       & 3.8 & 3.15 & 37\% \\
    \midrule
    \textbf{TPS (Ours)}            & 7.7 & 4.09 & 78\% \\
    Dominant Shuffle~\cite{zhao2024dominantshufflesimplepowerful} & 22.9 & 7.70 & 235\% \\
    \bottomrule
\end{tabular}
\end{table}

\paragraph{Results with CycleNet.}
Table~\ref{tab:cycle_etth1} reports results with CycleNet~\cite{lin2024cyclenetenhancingtimeseries}, a recent state-of-the-art forecasting model, on ETTh1. TPS achieves a \textbf{1.74\%} improvement over the second-best method, Dominant Shuffle, providing additional evidence that TPS can transfer effectively to strong forecasting backbones beyond the five main model families evaluated above.

\begin{table}[ht]
\centering
\footnotesize
\caption{
CycleNet on ETTh1 (MSE $\pm$ std). Best results are highlighted in bold, and second-best results are underlined.
}
\label{tab:cycle_etth1}
\setlength{\tabcolsep}{3pt}
\renewcommand{\arraystretch}{1.1}
\begin{tabular}{c|c|c|c|c||c}
\toprule
\textbf{Method} & \textbf{96} & \textbf{192} & \textbf{336} & \textbf{720} & \textbf{AVG} \\
\midrule
None & .368{\tiny$\pm$.002} & .407{\tiny$\pm$.004} & .406{\tiny$\pm$.003} & .446{\tiny$\pm$.003} & .407{\tiny$\pm$.003} \\
RobustTAD-m/p~\cite{gao2021robusttadrobusttimeseries} & .368{\tiny$\pm$.003} & .403{\tiny$\pm$.003} & .400{\tiny$\pm$.001} & .444{\tiny$\pm$.003} & .404{\tiny$\pm$.002} \\
FreqAdd~\cite{freqadd} & .368{\tiny$\pm$.005} & .406{\tiny$\pm$.005} & .400{\tiny$\pm$.003} & .441{\tiny$\pm$.007} & .404{\tiny$\pm$.004} \\
FreqPool~\cite{freqpool} & .402{\tiny$\pm$.001} & .413{\tiny$\pm$.002} & .408{\tiny$\pm$.001} & .447{\tiny$\pm$.002} & .418{\tiny$\pm$.001} \\
Upsample~\cite{upsample} & .377{\tiny$\pm$.001} & .412{\tiny$\pm$.003} & .402{\tiny$\pm$.002} & \secondres{.437{\tiny$\pm$.004}} & .407{\tiny$\pm$.002} \\
Freq-Mask/Mix~\cite{chen2023fraugfrequencydomainaugmentation} & \secondres{.366{\tiny$\pm$.001}} & .404{\tiny$\pm$.004} & .404{\tiny$\pm$.001} & .448{\tiny$\pm$.002} & .406{\tiny$\pm$.001} \\
Dominant Shuffle~\cite{zhao2024dominantshufflesimplepowerful} & \boldres{.364{\tiny$\pm$.003}} & \secondres{.401{\tiny$\pm$.001}} & \secondres{.400{\tiny$\pm$.002}} & .441{\tiny$\pm$.002} & \secondres{.402{\tiny$\pm$.003}} \\
\rowcolor{tpsblue} \textbf{TPS (Ours)} & .368{\tiny$\pm$.001} & \boldres{.399{\tiny$\pm$.003}} & \boldres{.387{\tiny$\pm$.004}} & \boldres{.424{\tiny$\pm$.002}} & \boldres{.395{\tiny$\pm$.003}} \\
\midrule
\textbf{Improvement} & --1.10\% & \boldres{0.50\%} & \boldres{3.25\%} & \boldres{2.97\%} & \boldres{1.74\%} \\
\bottomrule
\end{tabular}
\end{table}

\begin{figure*}[ht]
\begin{center}
    \centerline{\includegraphics[width=\textwidth]{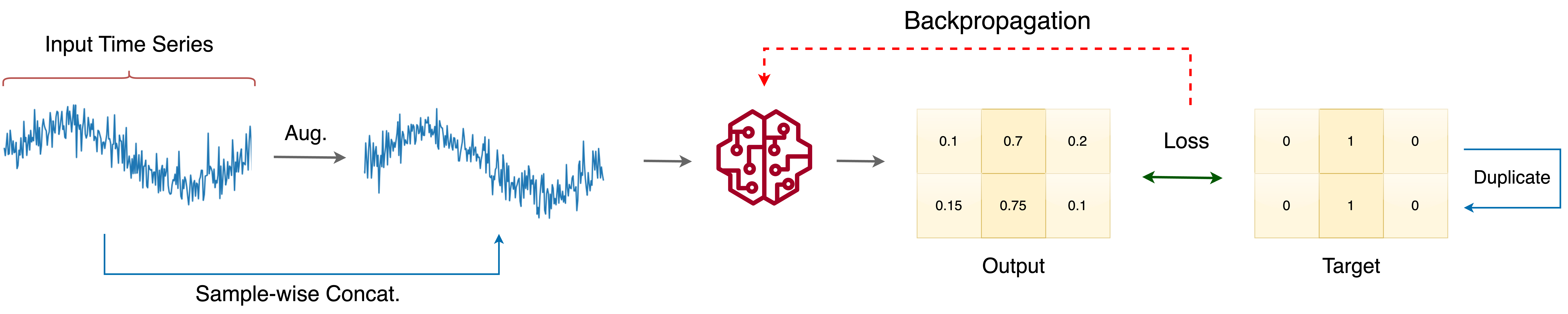}}
\caption{
Training pipeline for time series classification. Augmentation is applied to the input sequences only, while the corresponding labels are duplicated unchanged. The original and augmented samples are then concatenated along the sample dimension, effectively doubling the training batch, and the classifier is optimized using cross-entropy loss.
}
\Description{
Training pipeline for time series classification. The input time series is
augmented and concatenated with the original samples, while the class labels
are duplicated unchanged. A classifier produces predictions that are compared
with the labels using cross-entropy loss.
}
    \label{fig:tsc-frame}
\end{center}
\end{figure*}

\paragraph{TPS for Time Series Classification.}
We further evaluate TPS beyond forecasting on time series classification, using the widely adopted UCR~\cite{UCRArchive2018} and UEA~\cite{bagnall2018ueamultivariatetimeseries} repositories. For univariate classification, we train MiniRocket~\cite{dempster2021minirocket} on 30 datasets selected from the UCR archive, covering diverse categories. For multivariate classification, we use MultiRocket~\cite{tan2022multirocketmultiplepoolingoperators} on 10 datasets from the UEA archive, spanning a broad range of input dimensionalities, sequence lengths, and class counts. Original training data are split into 80\% training and 20\% validation; hyperparameters are selected on the validation set, and the model is then retrained on the full training set and evaluated on the provided test set. Full dataset statistics are provided in the supplementary material.

We compare TPS against transformation-based baselines (Jittering~\cite{Um_2017}, Rotation~\cite{10.1371/journal.pone.0254841}, Scaling~\cite{Um_2017}, Magnitude Warping~\cite{10.1371/journal.pone.0254841}, Window Slicing~\cite{leguennec:halshs-01357973}, Permutation and Random Permutation~\cite{Um_2017, Pan2020}, Time Warping~\cite{Um_2017, Park_2019}, and Window Warping~\cite{leguennec:halshs-01357973}) and pattern-based baselines (SPAWNER~\cite{s20010098}, wDBA~\cite{asd}, RGW and DGW~\cite{Um_2017}, and RGWs and DGWs~\cite{iwana2020timeseriesdataaugmentation}).

Extending TPS to classification requires two minor modifications. First, since classification operates only on the input sequence $\mathbf{X} \in \mathbb{R}^{N \times T \times C}$ (with no forecast horizon to concatenate), the temporal patching, variance-aware shuffling, and reconstruction blocks are applied directly to $\mathbf{X}$. Second, shuffling is performed at the sample level rather than the batch level. Class labels are unchanged and duplicated to match the augmented inputs, and the augmented samples are concatenated with the originals along the sample dimension, doubling the effective training set. The full classification training pipeline is illustrated in Figure~\ref{fig:tsc-frame}.

Table~\ref{tab:tsc_final} reports the results. TPS achieves the best average accuracy in both settings, improving over the second-best method by \textbf{0.50\%} on UCR and \textbf{1.10\%} on UEA, with 50\% and 60\% cumulative top-2 rankings, respectively. Full per-dataset results are available in the supplementary Excel file released with the code repository: \url{https://github.com/jafarbakhshaliyev/TPS/blob/main/results/results.xlsx}.

\begin{table}[ht]
\caption{
Accuracy of MiniRocket (univariate, 30 UCR datasets) and MultiRocket (multivariate, 10 UEA datasets) under different augmentation methods. For each dataset, accuracy is averaged over five runs; we report mean~$\pm$~std across datasets. Best is highlighted in bold, and second-best is underlined. 
}
\label{tab:tsc_final}
\centering
\footnotesize
\renewcommand{\arraystretch}{1.2}
\begin{tabular}{l|c|c}
\toprule
\textbf{Method} & \textbf{Univariate (UCR)} & \textbf{Multivariate (UEA)} \\
\midrule
None & 0.797 $\pm$ 0.0099 & 0.601 $\pm$ 0.0252 \\
Window Warping~\cite{leguennec:halshs-01357973} & 0.791 $\pm$ 0.0309 & \secondres{0.636 $\pm$ 0.0203} \\
Window Slicing~\cite{leguennec:halshs-01357973} & \secondres{0.800 $\pm$ 0.0125} & 0.631 $\pm$ 0.0310 \\
Jittering~\cite{Um_2017} & 0.786 $\pm$ 0.0140 & 0.631 $\pm$ 0.0258 \\
Scaling~\cite{Um_2017} & 0.793 $\pm$ 0.0133 & 0.608 $\pm$ 0.0232 \\
Permutation~\cite{Um_2017, Pan2020} & 0.796 $\pm$ 0.0164 & 0.603 $\pm$ 0.0228 \\
Rand. Permutation~\cite{Um_2017, Pan2020} & 0.789 $\pm$ 0.0144 & 0.619 $\pm$ 0.0181 \\
Time Warping~\cite{Um_2017, Park_2019} & 0.756 $\pm$ 0.0314 & 0.616 $\pm$ 0.0218 \\
Mag. Warping~\cite{10.1371/journal.pone.0254841} & 0.787 $\pm$ 0.0287 & 0.612 $\pm$ 0.0294 \\
Rotation~\cite{10.1371/journal.pone.0254841} & 0.793 $\pm$ 0.0153 & 0.607 $\pm$ 0.0363 \\
wDBA~\cite{asd} & 0.796 $\pm$ 0.0120 & 0.617 $\pm$ 0.0256 \\
RGW~\cite{Um_2017} & 0.790 $\pm$ 0.0130 & 0.631 $\pm$ 0.0241 \\
DGW~\cite{Um_2017} & 0.787 $\pm$ 0.0150 & 0.623 $\pm$ 0.0174 \\
SPAWNER~\cite{s20010098} & 0.785 $\pm$ 0.0122 & 0.630 $\pm$ 0.0258 \\
RGWs~\cite{iwana2020timeseriesdataaugmentation} & 0.799 $\pm$ 0.0128 & 0.633 $\pm$ 0.0192 \\
DGWs~\cite{iwana2020timeseriesdataaugmentation} & 0.797 $\pm$ 0.0132 & 0.621 $\pm$ 0.0285 \\
\rowcolor{tpsblue} \textbf{TPS (Ours)} & \boldres{0.804 $\pm$ 0.0098} & \boldres{0.643 $\pm$ 0.0253} \\
\midrule
\textbf{Improvement} & \boldres{0.50\%} & \boldres{1.10\%} \\
\textbf{\#Rank 1 (\%)} & 30.00 & 20.00 \\
\textbf{\#Rank 2 (\%)} & 20.00 & 40.00 \\
\bottomrule
\end{tabular}
\end{table}

\section{Discussion}

\paragraph{Limitations.}
TPS has several practical limitations. First, its performance depends on the perturbation strength controlled by $(p,s,\alpha)$; while TPS is generally robust under validation-based tuning, overly aggressive configurations can degrade performance on some datasets, as further reflected in the augmentation-size analysis reported in the supplementary material. Second, variance-aware selection provides a modest and configuration-dependent refinement: it affects partial-shuffling configurations but is inactive by construction when $\alpha=1$. Third, overlapping patch extraction and reconstruction introduce non-trivial training-time overhead compared with several frequency-domain augmentations.

\paragraph{Future Work.}
A natural direction for future work is to evaluate TPS across additional model families, such as Graph Neural Networks~\cite{yi2023fouriergnnrethinkingmultivariatetime} and Spiking Neural Networks~\cite{wu2025spikf, bakhshaliyev2026spikfgospikingfouriergraph}. Another promising direction is to refine the current variance-based ordering with more expressive patch-importance criteria, including channel-aware or weighted multivariate scoring. Third, TPS could be evaluated across a broader range of forecasting settings, including newer backbone families, pretrained time-series foundation models under lightweight adaptation protocols (e.g., frozen backbone with a fine-tuned prediction head), and standardized benchmark suites such as GIFT-Eval~\cite{aksu2024giftevalbenchmarkgeneraltime}.

\section{Conclusion}

In this work, we introduced Temporal Patch Shuffle (TPS), a simple and model-agnostic data augmentation method for time series forecasting that extracts overlapping temporal patches, applies controlled shuffling, and reconstructs the sequence by averaging overlapping regions. Across diverse forecasting models and datasets, TPS consistently outperforms existing augmentation techniques and provides a strong and practical augmentation strategy for time series forecasting. Our analyses indicate that preserving input--target coherence and using overlap-averaged reconstruction are central to controlling the temporal distortion introduced by patch reordering, while variance-aware selection provides a modest additional refinement under partial shuffling. Distributional analyses further show that TPS can introduce structured variation while remaining close to the temporal structure of the original data. In addition, TPS extends naturally beyond forecasting to time series classification, indicating its broader applicability within the time series domain.

\section*{GenAI Usage Disclosure}
ChatGPT 5.3 (OpenAI) was used solely for editing and proofreading assistance during the writing of this paper. All experimental code, method design, mathematical formulations, and analyses were produced entirely by the authors without the use of generative AI tools.

\bibliographystyle{ACM-Reference-Format}
\bibliography{example_paper}

\settopmatter{printacmref=false}

\clearpage
\settopmatter{printacmref=false}

\appendix

\section*{Extended Appendix (arXiv Version)}
The following material accompanies the CIKM 2026 proceedings paper
and is included in this arXiv version for completeness.

\section{Augmentation Baselines and Hyperparameters} \label{appendix:B}

Table~\ref{tab:aug_params} summarizes the hyperparameters used for all augmentation methods evaluated in this study, including our proposed method TPS. Most augmentation methods require only a small number of hyperparameters, and their recommended values are provided in the original papers. Nevertheless, we re-evaluated these settings to verify their effectiveness within our experimental setup.

For wDBA and MBB, we reproduce the methods based on the descriptions provided in their original papers~\cite{asd, mbb}. For RobustTAD, we follow the implementation details from~\cite{gao2021robusttadrobusttimeseries}: amplitude-based augmentation replaces segments with values sampled from a Gaussian distribution, while phase-based augmentation perturbs phase values using Gaussian noise. FreqAdd and FreqPool are implemented following~\cite{freqadd, freqpool}, where the former modifies the frequency components and the latter compresses the spectrum. For Upsample, we reproduce the method by selecting a consecutive segment and stretching it to match the length of the original time series~\cite{upsample}. For STAug, Freq-Mask/Mix, Wave-Mask/Mix, and Dominant Shuffle, we directly use the official implementations from their released codebases~\cite{zhang2023diversecoherentaugmentationtimeseries, chen2023fraugfrequencydomainaugmentation, arabi2024wavemaskmixexploringwaveletbasedaugmentations, zhao2024dominantshufflesimplepowerful}.

\begin{table}[ht]
\caption{Hyperparameters for the time series forecasting augmentation methods.}
\label{tab:aug_params}
\centering
\scriptsize
\renewcommand{\arraystretch}{1.2}
\setlength{\tabcolsep}{6pt}
\begin{tabular}{cl}
\toprule
\textbf{Method} & \textbf{Hyperparameters} \\
\midrule
wDBA~\cite{asd}                  & weighting, DTW constraints \\
MBB~\cite{mbb}                   & block size, STL period \\
RobustTAD-m/p~\cite{gao2021robusttadrobusttimeseries} & perturbation rate, \#segments, segment length \\
FreqAdd~\cite{freqadd}           & perturbation rate \\
FreqPool~\cite{freqpool}         & pool size \\
Upsample~\cite{upsample}         & subsequence length rate \\
STAug~\cite{zhang2023diversecoherentaugmentationtimeseries}          & mixup rate \\
FreqMask~\cite{chen2023fraugfrequencydomainaugmentation} & masking rate \\
FreqMix~\cite{chen2023fraugfrequencydomainaugmentation}  & mixing rate \\
Wave-Mask/Mix~\cite{arabi2024wavemaskmixexploringwaveletbasedaugmentations} & wavelet type, decomposition level, sampling rate \\
Dominant Shuffle~\cite{zhao2024dominantshufflesimplepowerful} & shuffle rate \\
\textbf{TPS (Ours)}              & patch length $(p)$, stride $(s)$, shuffle rate $(\alpha)$ \\
\bottomrule
\end{tabular}
\end{table}

\subsection{TPS Hyperparameter Selection}\label{appendix:tps_hparam}

We select $(p,s,\alpha)$ using a validation-based search over a predefined set of candidate values:
\begin{itemize}
\item Patch length:
\begin{equation*}
\begin{aligned}
p \in \{\,&16,\ 32,\ 48,\ 64,\ 72,\ 96,\ 120,\ 168,\ 192,\ 200,\ 220,\\
         &240,\ 280,\ 300,\ 340,\ 380,\ 400,\ 420,\ 440,\ 560,\ 700\,\}.
\end{aligned}
\end{equation*}
\item Stride: $s \in \{1,\ 2,\ 5,\ 8,\ 12,\ 16,\ 24,\ 32,\ 36,\ 96\}$.
\item Shuffle rate: $\alpha \in \{0.2,\ 0.5,\ 0.7,\ 0.8,\ 0.9,\ 1.0\}$.
\end{itemize}
To limit computational cost, we do not evaluate the full Cartesian product. Instead, we test a fixed set of approximately 20 candidate $(p,s,\alpha)$ combinations and select the one that minimizes validation MSE. The selected configuration is then evaluated once on the test set.

\section{Complementary Forecasting Results} \label{appendix:C}

\subsection{Long-Term Forecasting on the Common Seven-Dataset Subset}\label{appendix:C0}

Table~\ref{tab:mainresult_avg} reports the same long-term forecasting results as Table~2 in the main paper, but averaged over the common subset of seven datasets (excluding ECL and Traffic). This additional view is necessary because STAug cannot be evaluated on ECL and Traffic due to its high GPU memory requirements, consistent with the original STAug paper~\cite{zhang2023diversecoherentaugmentationtimeseries}.

TPS achieves substantial improvements of \textbf{2.98\%}, \textbf{5.83\%}, \textbf{3.16\%}, \textbf{2.26\%}, and \textbf{10.90\%} for TSMixer, DLinear, PatchTST, TiDE, and LightTS, respectively, while also obtaining a high number of wins across all models. We also observe that STAug underperforms several other augmentation baselines on this common subset.

\begin{table*}[ht]
\caption{
Long-term forecasting performance (MSE/MAE) averaged over seven datasets and four prediction lengths.
\#Wins is counted over 28 settings (7 datasets $\times$ 4 prediction lengths).
Best results are highlighted in bold and second-best results are underlined.
(ECL and Traffic are excluded to match the common subset used by all methods.)
}
\label{tab:mainresult_avg}
\centering
\scriptsize
\setlength{\tabcolsep}{5pt}
\renewcommand{\arraystretch}{1.2}
\begin{tabular}{c|cc|cc|cc|cc|cc}
\toprule
\textbf{Method} &
\multicolumn{2}{c|}{\textbf{TSMixer}} &
\multicolumn{2}{c|}{\textbf{DLinear}} &
\multicolumn{2}{c|}{\textbf{PatchTST}} &
\multicolumn{2}{c|}{\textbf{TiDE}} &
\multicolumn{2}{c}{\textbf{LightTS}} \\
& MSE & MAE & MSE & MAE & MSE & MAE & MSE & MAE & MSE & MAE \\
\midrule
None & 0.506 & 0.433 & 0.619 & 0.484 & 0.519 & 0.437 & 0.535 & 0.443 & 0.744 & 0.532 \\
wDBA~\cite{asd}  & 0.507 & \secondres{0.432} & 0.603 & 0.477 & 0.508 & 0.434 & 0.551 & 0.454 & 0.730 & 0.525 \\
MBB~\cite{mbb} & 0.514 & 0.435 & 0.611 & 0.480 & 0.507 & \secondres{0.432} & 0.550 & 0.451 & 0.746 & 0.533 \\
RobustTAD-m/p~\cite{gao2021robusttadrobusttimeseries} & 0.508 & 0.432 & 0.610 & 0.480 & 0.515 & 0.445 & 0.538 & 0.444 & 0.748 & 0.532 \\
FreqAdd~\cite{freqadd} & \secondres{0.504} & 0.433 & 0.628 & 0.489 & 0.518 & 0.438 & 0.533 & 0.443 & 0.739 & 0.530 \\
FreqPool~\cite{freqpool} & 0.520 & 0.439 & 0.593 & 0.475 & 0.529 & 0.442 & 0.551 & 0.452 & 0.709 & 0.515 \\
Upsample~\cite{upsample}  & 0.520 & 0.437 & \secondres{0.583} & \secondres{0.465} & 0.519 & 0.437 & 0.567 & 0.452 & \secondres{0.688} & \secondres{0.504} \\
STAug~\cite{zhang2023diversecoherentaugmentationtimeseries} & 0.540 & 0.451 & 0.733 & 0.538 & 0.548 & 0.453 & 0.630 & 0.497 & 0.855 & 0.583 \\
Freq-Mask/Mix~\cite{chen2023fraugfrequencydomainaugmentation} & 0.512 & 0.435 & 0.608 & 0.475 & 0.512 & 0.438 & 0.546 & 0.447 & 0.729 & 0.524 \\
Wave-Mask/Mix~\cite{arabi2024wavemaskmixexploringwaveletbasedaugmentations} & 0.509 & 0.433 & 0.615 & 0.481 & \secondres{0.507} & 0.436 & \secondres{0.531} & \secondres{0.441} & 0.739 & 0.527 \\
Dominant Shuffle~\cite{zhao2024dominantshufflesimplepowerful} & 0.517 & 0.437 & 0.615 & 0.477 & 0.521 & 0.442 & 0.548 & 0.446 & 0.725 & 0.518 \\
\rowcolor{tpsblue}
\textbf{TPS (Ours)}
& \boldres{0.489} & \boldres{0.424}
& \boldres{0.549} & \boldres{0.448}
& \boldres{0.491} & \boldres{0.424}
& \boldres{0.519} & \boldres{0.434}
& \boldres{0.613} & \boldres{0.478} \\
\midrule

\textbf{\#Wins (out of 28)} & 20 & 22 & 28 & 28 & 22 & 24 & 27 & 25 & 27 & 27 \\
\rowcolor{tpsblue}
\textbf{Improvement}
& \boldres{2.98\%} & \boldres{1.85\%}
& \boldres{5.83\%} & \boldres{3.66\%}
& \boldres{3.16\%} & \boldres{1.89\%}
& \boldres{2.26\%} & \boldres{1.59\%}
& \boldres{10.90\%} & \boldres{5.16\%} \\
\bottomrule
\end{tabular}
\end{table*}

\subsection{Long-Term Forecasting with Standard Deviations} \label{appendix:C1}

Tables~\ref{tab:appendix_longterm1} and~\ref{tab:appendix_longterm2} report long-term forecasting results on eight benchmark datasets (excluding ILI). We report mean MSE/MAE $\pm$ std, where the mean and standard deviation are first computed over five runs for each prediction length and then averaged over $\{96,192,336,720\}$.

TPS achieves the best performance in most settings, typically ranking first and occasionally second. The strongest competing methods vary by dataset, but are most often Freq-Mask/Mix, Wave-Mask/Mix, Dominant Shuffle, or Upsample. In addition to improved accuracy, TPS often reduces variability relative to training without augmentation, indicating that it can introduce useful diversity while keeping perturbations controlled.

Full results for each individual prediction length are also available in the supplementary Excel file released with the code: \url{https://github.com/jafarbakhshaliyev/TPS/blob/main/results/results.xlsx}.

\begin{table*}[ht]
\centering
\scriptsize
\setlength{\tabcolsep}{4pt}
\caption{
Long-term forecasting results averaged over prediction lengths $\{96,192,336,720\}$ on ETT-\{h1,h2,m1,m2\} for all five backbone models.
For each prediction length, we report mean $\pm$ std over 5 runs, then average these statistics over the four lengths.
Best results per dataset and model are highlighted in bold, and second-best results are underlined.
}
\label{tab:appendix_longterm1}
\resizebox{\textwidth}{!}{
\begin{tabular}{c|c|cc|cc|cc|cc}
\toprule
\textbf{} &
\textbf{Method} &
\multicolumn{2}{c|}{ETTh1} &
\multicolumn{2}{c|}{ETTh2} &
\multicolumn{2}{c}{ETTm1} &
\multicolumn{2}{c}{ETTm2} \\
 & & MSE & MAE & MSE & MAE & MSE & MAE & MSE & MAE \\
\midrule
\multirow{11}{*}{\rotatebox{90}{\textbf{TSMixer}}}
 & None
      & 0.413 $\pm$ 0.0023 & 0.434 $\pm$ 0.0020
      & 0.342 $\pm$ 0.0023 & 0.394 $\pm$ 0.0017
      & 0.351 $\pm$ 0.0010 & 0.378 $\pm$ 0.0008
      & 0.262 $\pm$ 0.0017 & 0.320 $\pm$ 0.0014 \\
 &  wDBA~\cite{asd}
      & 0.418 $\pm$ 0.0028 & 0.437 $\pm$ 0.0026
      & 0.344 $\pm$ 0.0023 & 0.394 $\pm$ 0.0019
      & 0.352 $\pm$ 0.0009 & 0.378 $\pm$ 0.0008
      & 0.261 $\pm$ 0.0011 & \secondres{0.319 $\pm$ 0.0011}\\
 & MBB~\cite{mbb}
      & 0.412 $\pm$ 0.0026 & 0.437 $\pm$ 0.0023
      & 0.345 $\pm$ 0.0046 & 0.395 $\pm$ 0.0034
      & 0.353 $\pm$ 0.0014 & 0.379 $\pm$ 0.0013
      & 0.264 $\pm$ 0.0017 & 0.321 $\pm$ 0.0011 \\
 & RobustTAD-m/p~\cite{gao2021robusttadrobusttimeseries}
      & 0.416 $\pm$ 0.0018 & 0.434 $\pm$ 0.0016
      & 0.343 $\pm$ 0.0032 & 0.394 $\pm$ 0.0023
      & 0.350 $\pm$ 0.0008 & 0.377 $\pm$ 0.0008
      & 0.262 $\pm$ 0.0037 & 0.319 $\pm$ 0.0021 \\
 & FreqAdd~\cite{freqadd}
      & 0.413 $\pm$ 0.0026 & 0.432 $\pm$ 0.0024
      & 0.343 $\pm$ 0.0025 & 0.393 $\pm$ 0.0016
      & 0.351 $\pm$ 0.0010 & 0.377 $\pm$ 0.0008
      & 0.263 $\pm$ 0.0062 & 0.320 $\pm$ 0.0035 \\
 & FreqPool~\cite{freqpool}
      & 0.431 $\pm$ 0.0055 & 0.446 $\pm$ 0.0043
      & 0.351 $\pm$ 0.0031 & 0.399 $\pm$ 0.0020
      & 0.359 $\pm$ 0.0010 & 0.382 $\pm$ 0.0010
      & 0.264 $\pm$ 0.0015 & 0.322 $\pm$ 0.0013 \\
 & Upsample~\cite{upsample}
      & 0.416 $\pm$ 0.0023 & 0.434 $\pm$ 0.0023
      & 0.343 $\pm$ 0.0019 & \secondres{0.392 $\pm$ 0.0014}
      & 0.378 $\pm$ 0.0013 & 0.396 $\pm$ 0.0009
      & 0.263 $\pm$ 0.0016 & 0.320 $\pm$ 0.0010 \\
 & STAug~\cite{zhang2023diversecoherentaugmentationtimeseries}
      & 0.411 $\pm$ 0.0020 & 0.432 $\pm$ 0.0019
      & 0.393 $\pm$ 0.0067 & 0.427 $\pm$ 0.0032
      & 0.352 $\pm$ 0.0017 & 0.377 $\pm$ 0.0015
      & 0.339 $\pm$ 0.0055 & 0.373 $\pm$ 0.0021 \\
 & Freq-Mask/Mix~\cite{chen2023fraugfrequencydomainaugmentation}
      & 0.416 $\pm$ 0.0035 & 0.434 $\pm$ 0.0027
      & 0.344 $\pm$ 0.0019 & 0.394 $\pm$ 0.0015
      & \secondres{0.348 $\pm$ 0.0010} & \secondres{0.376 $\pm$ 0.0006}
      & \secondres{0.259 $\pm$ 0.0030} & 0.319 $\pm$ 0.0018 \\
 & Wave-Mask/Mix~\cite{arabi2024wavemaskmixexploringwaveletbasedaugmentations}
      & \secondres{0.410 $\pm$ 0.0021} & \secondres{0.431 $\pm$ 0.0021}
      & \secondres{0.342 $\pm$ 0.0022} & 0.393 $\pm$ 0.0016
      & 0.352 $\pm$ 0.0013 & 0.379 $\pm$ 0.0009
      & 0.261 $\pm$ 0.0015 & 0.320 $\pm$ 0.0009 \\
 & Dominant Shuffle~\cite{zhao2024dominantshufflesimplepowerful}
      & 0.416 $\pm$ 0.0019 & 0.434 $\pm$ 0.0023
      & 0.345 $\pm$ 0.0027 & 0.395 $\pm$ 0.0020
      & 0.353 $\pm$ 0.0012 & 0.380 $\pm$ 0.0009
      & 0.261 $\pm$ 0.0017 & 0.320 $\pm$ 0.0007 \\
\rowcolor{tpsblue}
 &   \textbf{TPS (Ours)}
      & \boldres{0.402 $\pm$ 0.0018} & \boldres{0.425 $\pm$ 0.0019}
      & \boldres{0.339 $\pm$ 0.0021} & \boldres{0.390 $\pm$ 0.0015}
      & \boldres{0.346 $\pm$ 0.0010} & \boldres{0.373 $\pm$ 0.0007}
      & \boldres{0.257 $\pm$ 0.0016} & \boldres{0.317 $\pm$ 0.0008} \\
\midrule
\multirow{11}{*}{\rotatebox{90}{\textbf{DLinear}}}
 & None & 0.438 $\pm$ 0.0194 & 0.449 $\pm$ 0.0162 & 0.464 $\pm$ 0.0099 & 0.462 $\pm$ 0.0053 & 0.361 $\pm$ 0.0010 & 0.383 $\pm$ 0.0015 & 0.276 $\pm$ 0.0099 & 0.339 $\pm$ 0.0093 \\
 & wDBA & 0.432 $\pm$ 0.0126 & 0.443 $\pm$ 0.0126 & 0.434 $\pm$ 0.0383 & 0.444 $\pm$ 0.0175 & 0.360 $\pm$ 0.0011 & \secondres{0.381 $\pm$ 0.0016} & 0.270 $\pm$ 0.0050 & 0.334 $\pm$ 0.0055 \\
 & MBB & 0.429 $\pm$ 0.0114 & 0.444 $\pm$ 0.0111 & 0.449 $\pm$ 0.0324 & 0.454 $\pm$ 0.0152 & 0.360 $\pm$ 0.0009 & 0.382 $\pm$ 0.0015 & 0.280 $\pm$ 0.0080 & 0.343 $\pm$ 0.0072 \\
 & RobustTAD-m/p & 0.433 $\pm$ 0.0155 & 0.444 $\pm$ 0.0129 & 0.432 $\pm$ 0.0295 & 0.447 $\pm$ 0.0151 & 0.361 $\pm$ 0.0012 & 0.383 $\pm$ 0.0017 & 0.280 $\pm$ 0.0088 & 0.342 $\pm$ 0.0071 \\
 & FreqAdd & 0.432 $\pm$ 0.0128 & 0.446 $\pm$ 0.0120 & 0.471 $\pm$ 0.0195 & 0.466 $\pm$ 0.0090 & 0.364 $\pm$ 0.0009 & 0.386 $\pm$ 0.0012 & 0.285 $\pm$ 0.0097 & 0.347 $\pm$ 0.0085 \\
 & FreqPool & 0.450 $\pm$ 0.0231 & 0.455 $\pm$ 0.0166 & 0.422 $\pm$ 0.0177 & 0.439 $\pm$ 0.0104 & 0.380 $\pm$ 0.0008 & 0.399 $\pm$ 0.0011 & 0.273 $\pm$ 0.0111 & 0.335 $\pm$ 0.0075 \\
 & Upsample & 0.436 $\pm$ 0.0059 & 0.445 $\pm$ 0.0055 & \secondres{0.391 $\pm$ 0.0178} & \secondres{0.423 $\pm$ 0.0098} & 0.387 $\pm$ 0.0007 & 0.406 $\pm$ 0.0015 & \secondres{0.265 $\pm$ 0.0048} & \secondres{0.328 $\pm$ 0.0048} \\
 & STAug
    & 0.442 $\pm$ 0.0345 & 0.450 $\pm$ 0.0226
    & 1.011 $\pm$ 0.1382 & 0.678 $\pm$ 0.0543
    & \secondres{0.359 $\pm$ 0.0016} & 0.383 $\pm$ 0.0023
    & 0.442 $\pm$ 0.0212 & 0.416 $\pm$ 0.0131 \\
 & Freq-Mask/Mix & 0.422 $\pm$ 0.0063 & 0.436 $\pm$ 0.0066 & 0.422 $\pm$ 0.0246 & 0.440 $\pm$ 0.0133 & 0.360 $\pm$ 0.0009 & 0.383 $\pm$ 0.0013 & 0.271 $\pm$ 0.0037 & 0.336 $\pm$ 0.0044 \\
 & Wave-Mask/Mix & 0.426 $\pm$ 0.0058 & 0.441 $\pm$ 0.0056 & 0.454 $\pm$ 0.0228 & 0.456 $\pm$ 0.0118 & 0.361 $\pm$ 0.0014 & 0.383 $\pm$ 0.0019 & 0.275 $\pm$ 0.0108 & 0.338 $\pm$ 0.0096 \\
 & Dominant Shuffle & \secondres{0.420 $\pm$ 0.0037} & \secondres{0.434 $\pm$ 0.0044} & 0.409 $\pm$ 0.0104 & 0.435 $\pm$ 0.0066 & 0.360 $\pm$ 0.0014 & 0.383 $\pm$ 0.0015 & 0.271 $\pm$ 0.0089 & 0.335 $\pm$ 0.0095 \\
 \rowcolor{tpsblue}
 & \textbf{TPS (Ours)} & \boldres{0.410 $\pm$ 0.0036} & \boldres{0.425 $\pm$ 0.0031} &
         \boldres{0.369 $\pm$ 0.0056} & \boldres{0.408 $\pm$ 0.0039} &
         \boldres{0.354 $\pm$ 0.0006} & \boldres{0.377 $\pm$ 0.0006} &
         \boldres{0.261 $\pm$ 0.0018} & \boldres{0.324 $\pm$ 0.0027} \\
\midrule
\multirow{11}{*}{\rotatebox{90}{\textbf{PatchTST}}}
 & None & 0.414 $\pm$ 0.0026 & 0.428 $\pm$ 0.0023 & 0.331 $\pm$ 0.0016 & 0.380 $\pm$ 0.0020 & 0.354 $\pm$ 0.0030 & 0.383 $\pm$ 0.0019 & 0.258 $\pm$ 0.0018 & 0.316 $\pm$ 0.0013 \\
 & wDBA & 0.417 $\pm$ 0.0047 & 0.431 $\pm$ 0.0035 & 0.335 $\pm$ 0.0009 & 0.382 $\pm$ 0.0011 & 0.352 $\pm$ 0.0019 & 0.381 $\pm$ 0.0010 & 0.261 $\pm$ 0.0011 & 0.318 $\pm$ 0.0010 \\
 & MBB & 0.422 $\pm$ 0.0057 & 0.433 $\pm$ 0.0033 & 0.332 $\pm$ 0.0011 & 0.381 $\pm$ 0.0014 & 0.353 $\pm$ 0.0019 & 0.383 $\pm$ 0.0008 & 0.258 $\pm$ 0.0004 & 0.316 $\pm$ 0.0007 \\
 & RobustTAD-m/p & 0.423 $\pm$ 0.0116 & 0.434 $\pm$ 0.0058 & 0.331 $\pm$ 0.0009 & 0.380 $\pm$ 0.0059 & 0.354 $\pm$ 0.0025 & 0.383 $\pm$ 0.0018 & 0.258 $\pm$ 0.0020 & 0.317 $\pm$ 0.0016 \\
 & FreqAdd & 0.413 $\pm$ 0.0022 & 0.429 $\pm$ 0.0018 & 0.333 $\pm$ 0.0013 & 0.381 $\pm$ 0.0019 & 0.355 $\pm$ 0.0041 & 0.384 $\pm$ 0.0014 & 0.261 $\pm$ 0.0022 & 0.317 $\pm$ 0.0016 \\
 & FreqPool & 0.423 $\pm$ 0.0039 & 0.436 $\pm$ 0.0029 & 0.333 $\pm$ 0.0012 & 0.381 $\pm$ 0.0008 & 0.351 $\pm$ 0.0024 & 0.381 $\pm$ 0.0015 & \secondres{0.256 $\pm$ 0.0018} & \secondres{0.315 $\pm$ 0.0008} \\
 & Upsample & 0.419 $\pm$ 0.0047 & 0.433 $\pm$ 0.0029 & 0.335 $\pm$ 0.0010 & 0.382 $\pm$ 0.0011 & 0.356 $\pm$ 0.0037 & 0.387 $\pm$ 0.0021 & 0.258 $\pm$ 0.0013 & 0.316 $\pm$ 0.0006 \\
 & STAug
    & 0.419 $\pm$ 0.0074 & 0.429 $\pm$ 0.0033
    & 0.416 $\pm$ 0.0115 & 0.428 $\pm$ 0.0047
    & \secondres{0.348 $\pm$ 0.0025} & \secondres{0.379 $\pm$ 0.0016}
    & 0.275 $\pm$ 0.0036 & 0.325 $\pm$ 0.0017 \\
 & Freq-Mask/Mix & 0.412 $\pm$ 0.0024 & 0.427 $\pm$ 0.0023 & 0.329 $\pm$ 0.0012 & 0.380 $\pm$ 0.0009 & 0.351 $\pm$ 0.0019 & 0.382 $\pm$ 0.0011 & 0.261 $\pm$ 0.0038 & 0.318 $\pm$ 0.0025 \\
 & Wave-Mask/Mix & 0.412 $\pm$ 0.0029 & 0.428 $\pm$ 0.0021 & 0.330 $\pm$ 0.0017 & \secondres{0.379 $\pm$ 0.0018} & 0.352 $\pm$ 0.0027 & 0.381 $\pm$ 0.0018 & 0.258 $\pm$ 0.0012 & 0.316 $\pm$ 0.0012 \\
 & Dominant Shuffle & \secondres{0.408 $\pm$ 0.0015} & \secondres{0.424 $\pm$ 0.0010} & \secondres{0.329 $\pm$ 0.0009} & 0.382 $\pm$ 0.0012 & 0.354 $\pm$ 0.0022 & 0.383 $\pm$ 0.0017 & 0.258 $\pm$ 0.0014 & 0.316 $\pm$ 0.0013 \\
 \rowcolor{tpsblue} &  \textbf{TPS (Ours)} & \boldres{0.401 $\pm$ 0.0021} & \boldres{0.419 $\pm$ 0.0021} &
         \boldres{0.326 $\pm$ 0.0013} & \boldres{0.378 $\pm$ 0.0008} &
         \boldres{0.345 $\pm$ 0.0019} & \boldres{0.377 $\pm$ 0.0009} &
         \boldres{0.256 $\pm$ 0.0013} & \boldres{0.315 $\pm$ 0.0009} \\
\midrule
\multirow{11}{*}{\rotatebox{90}{\textbf{TiDE}}}
 & None & 0.417 $\pm$ 0.0012 & 0.432 $\pm$ 0.0011 & 0.316 $\pm$ 0.0018 & 0.375 $\pm$ 0.0011 & 0.359 $\pm$ 0.0036 & 0.381 $\pm$ 0.0032 & 0.250 $\pm$ 0.0009 & 0.313 $\pm$ 0.0005 \\
 & wDBA & 0.563 $\pm$ 0.0055 & 0.518 $\pm$ 0.0023 & 0.321 $\pm$ 0.0009 & 0.378 $\pm$ 0.0006 & 0.360 $\pm$ 0.0027 & 0.382 $\pm$ 0.0025 & 0.250 $\pm$ 0.0006 & 0.313 $\pm$ 0.0009 \\
 & MBB & 0.556 $\pm$ 0.0027 & 0.513 $\pm$ 0.0011 & \secondres{0.311 $\pm$ 0.0006} & \secondres{0.371 $\pm$ 0.0004} & 0.358 $\pm$ 0.0025 & 0.381 $\pm$ 0.0020 & 0.250 $\pm$ 0.0003 & \secondres{0.312 $\pm$ 0.0004} \\
 & RobustTAD-m/p & 0.418 $\pm$ 0.0019 & 0.433 $\pm$ 0.0012 & 0.316 $\pm$ 0.0008 & 0.375 $\pm$ 0.0005 & 0.357 $\pm$ 0.0008 & 0.380 $\pm$ 0.0011 & 0.251 $\pm$ 0.0009 & 0.313 $\pm$ 0.0005 \\
 & FreqAdd & 0.406 $\pm$ 0.0007 & 0.428 $\pm$ 0.0006 & 0.312 $\pm$ 0.0007 & 0.371 $\pm$ 0.0005 & 0.361 $\pm$ 0.0022 & 0.383 $\pm$ 0.0018 & 0.252 $\pm$ 0.0017 & 0.315 $\pm$ 0.0009 \\
 & FreqPool & 0.431 $\pm$ 0.0032 & 0.442 $\pm$ 0.0026 & 0.322 $\pm$ 0.0010 & 0.380 $\pm$ 0.0006 & 0.377 $\pm$ 0.0021 & 0.395 $\pm$ 0.0016 & 0.253 $\pm$ 0.0009 & 0.316 $\pm$ 0.0004 \\
 & Upsample & 0.423 $\pm$ 0.0008 & 0.439 $\pm$ 0.0006 & 0.328 $\pm$ 0.0043 & 0.381 $\pm$ 0.0023 & 0.390 $\pm$ 0.0042 & 0.402 $\pm$ 0.0023 & 0.252 $\pm$ 0.0005 & 0.314 $\pm$ 0.0003 \\
 & STAug
    & 0.533 $\pm$ 0.0038 & 0.514 $\pm$ 0.0018
    & 0.571 $\pm$ 0.0751 & 0.521 $\pm$ 0.0382
    & 0.356 $\pm$ 0.0016 & 0.379 $\pm$ 0.0017
    & 0.421 $\pm$ 0.0099 & 0.396 $\pm$ 0.0037 \\
 & Freq-Mask/Mix & 0.421 $\pm$ 0.0017 & 0.436 $\pm$ 0.0011 & 0.317 $\pm$ 0.0008 & 0.377 $\pm$ 0.0007 & 0.355 $\pm$ 0.0011 & \secondres{0.379 $\pm$ 0.0009} & 0.253 $\pm$ 0.0012 & 0.317 $\pm$ 0.0007 \\
 & Wave-Mask/Mix & \secondres{0.402 $\pm$ 0.0005} & \secondres{0.426 $\pm$ 0.0006} & 0.312 $\pm$ 0.0009 & 0.372 $\pm$ 0.0007 & 0.357 $\pm$ 0.0014 & 0.380 $\pm$ 0.0011 & \secondres{0.249 $\pm$ 0.0003} & 0.312 $\pm$ 0.0005 \\
 & Dominant Shuffle & 0.412 $\pm$ 0.0010 & 0.429 $\pm$ 0.0007 & 0.312 $\pm$ 0.0010 & 0.374 $\pm$ 0.0005 & \secondres{0.353 $\pm$ 0.0020} & 0.379 $\pm$ 0.0018 & 0.255 $\pm$ 0.0007 & 0.317 $\pm$ 0.0006 \\
 \rowcolor{tpsblue}
 &   \textbf{TPS (Ours)} & \boldres{0.387 $\pm$ 0.0010} & \boldres{0.415 $\pm$ 0.0008} &
         \boldres{0.308 $\pm$ 0.0005} & \boldres{0.370 $\pm$ 0.0004} &
         \boldres{0.347 $\pm$ 0.0019} & \boldres{0.373 $\pm$ 0.0017} &
         \boldres{0.248 $\pm$ 0.0003} & \boldres{0.311 $\pm$ 0.0003} \\
\midrule
\multirow{11}{*}{\rotatebox{90}{\textbf{LightTS}}}
 & None & 0.462 $\pm$ 0.0051 & 0.473 $\pm$ 0.0030 & 0.611 $\pm$ 0.0152 & 0.540 $\pm$ 0.0089 & 0.380 $\pm$ 0.0025 & 0.401 $\pm$ 0.0021 & 0.301 $\pm$ 0.0052 & 0.363 $\pm$ 0.0050 \\
 & wDBA & 0.462 $\pm$ 0.0031 & 0.473 $\pm$ 0.0025 & 0.588 $\pm$ 0.0115 & 0.529 $\pm$ 0.0059 & 0.381 $\pm$ 0.0051 & 0.401 $\pm$ 0.0035 & 0.290 $\pm$ 0.0050 & 0.352 $\pm$ 0.0042 \\
 & MBB & 0.454 $\pm$ 0.0026 & 0.469 $\pm$ 0.0023 & 0.612 $\pm$ 0.0157 & 0.541 $\pm$ 0.0071 & 0.382 $\pm$ 0.0030 & 0.404 $\pm$ 0.0027 & 0.306 $\pm$ 0.0081 & 0.367 $\pm$ 0.0076 \\
 & RobustTAD-m/p & 0.462 $\pm$ 0.0023 & 0.473 $\pm$ 0.0015 & 0.597 $\pm$ 0.0127 & 0.535 $\pm$ 0.0074 & 0.382 $\pm$ 0.0032 & 0.402 $\pm$ 0.0029 & 0.299 $\pm$ 0.0021 & 0.360 $\pm$ 0.0025 \\
 & FreqAdd & 0.454 $\pm$ 0.0032 & 0.468 $\pm$ 0.0031 & 0.615 $\pm$ 0.0129 & 0.541 $\pm$ 0.0061 & 0.378 $\pm$ 0.0028 & 0.400 $\pm$ 0.0025 & 0.300 $\pm$ 0.0099 & 0.361 $\pm$ 0.0091 \\
 & FreqPool & 0.482 $\pm$ 0.0057 & 0.487 $\pm$ 0.0040 & 0.543 $\pm$ 0.0155 & 0.516 $\pm$ 0.0089 & 0.392 $\pm$ 0.0027 & 0.410 $\pm$ 0.0027 & 0.297 $\pm$ 0.0060 & 0.358 $\pm$ 0.0037 \\
 & Upsample & 0.468 $\pm$ 0.0030 & 0.478 $\pm$ 0.0020 & \secondres{0.446 $\pm$ 0.0066} & \secondres{0.464 $\pm$ 0.0034} & 0.402 $\pm$ 0.0054 & 0.415 $\pm$ 0.0038 & \secondres{0.283 $\pm$ 0.0018} & \secondres{0.347 $\pm$ 0.0027} \\
 & STAug
    & 0.458 $\pm$ 0.0034 & 0.470 $\pm$ 0.0026
    & 1.248 $\pm$ 0.0336 & 0.782 $\pm$ 0.0119
    & \secondres{0.363 $\pm$ 0.0014} & \secondres{0.388 $\pm$ 0.0014}
    & 0.546 $\pm$ 0.0165 & 0.486 $\pm$ 0.0083 \\
 & Freq-Mask/Mix & 0.461 $\pm$ 0.0028 & 0.473 $\pm$ 0.0018 & 0.558 $\pm$ 0.0110 & 0.518 $\pm$ 0.0063 & 0.373 $\pm$ 0.0016 & 0.398 $\pm$ 0.0014 & 0.297 $\pm$ 0.0119 & 0.360 $\pm$ 0.0118 \\
 & Wave-Mask/Mix & \secondres{0.448 $\pm$ 0.0031} & 0.466 $\pm$ 0.0021 & 0.611 $\pm$ 0.0086 & 0.539 $\pm$ 0.0047 & 0.377 $\pm$ 0.0018 & 0.400 $\pm$ 0.0024 & 0.299 $\pm$ 0.0075 & 0.360 $\pm$ 0.0074 \\
 & Dominant Shuffle & 0.449 $\pm$ 0.0036 & \secondres{0.464 $\pm$ 0.0028} & 0.511 $\pm$ 0.0101 & 0.496 $\pm$ 0.0058 & 0.372 $\pm$ 0.0038 & 0.398 $\pm$ 0.0031 & 0.290 $\pm$ 0.0035 & 0.353 $\pm$ 0.0043 \\
\rowcolor{tpsblue}
 &  \textbf{TPS (Ours)} & \boldres{0.431 $\pm$ 0.0007} & \boldres{0.452 $\pm$ 0.0005} &
         \boldres{0.418 $\pm$ 0.0031} & \boldres{0.447 $\pm$ 0.0019} &
         \boldres{0.356 $\pm$ 0.0015} & \boldres{0.382 $\pm$ 0.0015} &
         \boldres{0.277 $\pm$ 0.0027} & \boldres{0.342 $\pm$ 0.0029} \\
\bottomrule
\end{tabular}}

\end{table*}

\begin{table*}[ht]
\centering
\scriptsize
\setlength{\tabcolsep}{4pt}
\caption{
Long-term forecasting results averaged over prediction lengths $\{96,192,336,720\}$ on Exchange, Weather, ECL, and Traffic for all five backbone models.
For each prediction length, we report mean $\pm$ std over 5 runs, then average these statistics over the four lengths.
Best results per dataset and model are highlighted in bold, and second-best results are underlined.
STAug is not reported on ECL and Traffic due to GPU memory limitations.
}
\label{tab:appendix_longterm2}
\resizebox{\textwidth}{!}{
\begin{tabular}{c|c|cc|cc|cc|cc}
\toprule
\textbf{} &
\textbf{Method} &
\multicolumn{2}{c|}{Exchange} &
\multicolumn{2}{c|}{Weather} &
\multicolumn{2}{c|}{ECL} &
\multicolumn{2}{c}{Traffic} \\
 & & MSE & MAE & MSE & MAE & MSE & MAE & MSE & MAE \\
\midrule
\multirow{11}{*}{\rotatebox{90}{\textbf{TSMixer}}}
 & None &
    0.417 $\pm$ 0.0074 & 0.436 $\pm$ 0.0038 &
    0.224 $\pm$ 0.0016 & 0.263 $\pm$ 0.0014 &
    0.170 $\pm$ 0.0003 & 0.269 $\pm$ 0.0004 &
    0.436 $\pm$ 0.0007 & 0.327 $\pm$ 0.0009 \\
 & wDBA~\cite{asd} &
    0.406 $\pm$ 0.0091 & 0.430 $\pm$ 0.0050 &
    0.222 $\pm$ 0.0006 & 0.261 $\pm$ 0.0006 &
    0.173 $\pm$ 0.0002 & 0.275 $\pm$ 0.0002 &
    0.437 $\pm$ 0.0010 & 0.326 $\pm$ 0.0013 \\
 & MBB~\cite{mbb} &
    0.408 $\pm$ 0.0074 & 0.431 $\pm$ 0.0038 &
    0.224 $\pm$ 0.0006 & 0.263 $\pm$ 0.0006 &
    0.171 $\pm$ 0.0001 & 0.271 $\pm$ 0.0002 &
    0.438 $\pm$ 0.0069 & 0.325 $\pm$ 0.0077 \\
 & RobustTAD-m/p~\cite{gao2021robusttadrobusttimeseries} &
    0.409 $\pm$ 0.0042 & 0.431 $\pm$ 0.0028 &
    0.223 $\pm$ 0.0010 & 0.262 $\pm$ 0.0010 &
    0.168 $\pm$ 0.0003 & 0.268 $\pm$ 0.0005 &
    \secondres{0.431 $\pm$ 0.0004} & \secondres{0.317 $\pm$ 0.0008} \\
 & FreqAdd~\cite{freqadd} &
    0.407 $\pm$ 0.0071 & 0.436 $\pm$ 0.0035 &
    0.226 $\pm$ 0.0007 & 0.265 $\pm$ 0.0006 &
    \secondres{0.168 $\pm$ 0.0002} & \secondres{0.268 $\pm$ 0.0002} &
    0.432 $\pm$ 0.0006 & 0.319 $\pm$ 0.0009 \\
 & FreqPool~\cite{freqpool} &
    0.418 $\pm$ 0.0032 & 0.434 $\pm$ 0.0021 &
    0.224 $\pm$ 0.0009 & 0.264 $\pm$ 0.0011 &
    0.176 $\pm$ 0.0004 & 0.277 $\pm$ 0.0004 &
    0.442 $\pm$ 0.0009 & 0.325 $\pm$ 0.0005 \\
 & Upsample~\cite{upsample} &
    \secondres{0.401 $\pm$ 0.0053} & \secondres{0.425 $\pm$ 0.0032} &
    \boldres{0.217 $\pm$ 0.0008} & \boldres{0.259 $\pm$ 0.0009} &
    0.187 $\pm$ 0.0003 & 0.286 $\pm$ 0.0004 &
    0.463 $\pm$ 0.0008 & 0.346 $\pm$ 0.0008 \\
 & STAug~\cite{zhang2023diversecoherentaugmentationtimeseries} &
    0.413 $\pm$ 0.0061 & 0.434 $\pm$ 0.0037 &
    0.241 $\pm$ 0.0046 & 0.285 $\pm$ 0.0058 &
    - & - &
    - & - \\
 & Freq-Mask/Mix~\cite{chen2023fraugfrequencydomainaugmentation} &
    0.431 $\pm$ 0.0068 & 0.444 $\pm$ 0.0042 &
    \secondres{0.222 $\pm$ 0.0006} & 0.262 $\pm$ 0.0007 &
    0.174 $\pm$ 0.0002 & 0.276 $\pm$ 0.0003 &
    0.450 $\pm$ 0.0008 & 0.323 $\pm$ 0.0004 \\
 & Wave-Mask/Mix~\cite{arabi2024wavemaskmixexploringwaveletbasedaugmentations} &
    0.413 $\pm$ 0.0073 & 0.434 $\pm$ 0.0036 &
    0.225 $\pm$ 0.0007 & 0.265 $\pm$ 0.0010 &
    0.169 $\pm$ 0.0002 & 0.269 $\pm$ 0.0003 &
    0.434 $\pm$ 0.0008 & 0.323 $\pm$ 0.0007 \\
 & Dominant Shuffle~\cite{zhao2024dominantshufflesimplepowerful} &
    0.401 $\pm$ 0.0057 & 0.431 $\pm$ 0.0036 &
    0.225 $\pm$ 0.0010 & 0.264 $\pm$ 0.0009 &
    0.175 $\pm$ 0.0002 & 0.276 $\pm$ 0.0003 &
    0.444 $\pm$ 0.0006 & 0.326 $\pm$ 0.0007 \\
    
  \rowcolor{tpsblue} & \textbf{TPS (Ours)} &
    \boldres{0.391 $\pm$ 0.0067} & \boldres{0.421 $\pm$ 0.0036} &
    0.222 $\pm$ 0.0010 & \secondres{0.260 $\pm$ 0.0007} &
    \boldres{0.168 $\pm$ 0.0002} & \boldres{0.267 $\pm$ 0.0002} &
    \boldres{0.428 $\pm$ 0.0004} & \boldres{0.313 $\pm$ 0.0008} \\
\midrule
\multirow{11}{*}{\rotatebox{90}{\textbf{DLinear}}}
 & None &
    0.381 $\pm$ 0.0478 & 0.421 $\pm$ 0.0199 &
    \secondres{0.245 $\pm$ 0.0004} & 0.298 $\pm$ 0.0009 &
    0.166 $\pm$ 0.0000 & 0.264 $\pm$ 0.0002 &
    0.434 $\pm$ 0.0000 & 0.295 $\pm$ 0.0000 \\
 & wDBA &
    0.386 $\pm$ 0.0602 & 0.422 $\pm$ 0.0311 &
    0.245 $\pm$ 0.0031 & 0.296 $\pm$ 0.0062 &
    0.173 $\pm$ 0.0000 & \secondres{0.263 $\pm$ 0.0004} &
    0.434 $\pm$ 0.0000 & \secondres{0.295 $\pm$ 0.0000} \\
 & MBB &
    0.383 $\pm$ 0.0414 & 0.419 $\pm$ 0.0164 &
    0.246 $\pm$ 0.0025 & 0.299 $\pm$ 0.0056 &
    0.173 $\pm$ 0.0001 & 0.273 $\pm$ 0.0001 &
    0.450 $\pm$ 0.0000 & 0.315 $\pm$ 0.0002 \\
 & RobustTAD-m/p &
    0.363 $\pm$ 0.0539 & 0.412 $\pm$ 0.0234 &
    0.245 $\pm$ 0.0005 & 0.297 $\pm$ 0.0008 &
    0.167 $\pm$ 0.0000 & 0.264 $\pm$ 0.0001 &
    0.434 $\pm$ 0.0000 & 0.296 $\pm$ 0.0000 \\
 & FreqAdd &
    0.385 $\pm$ 0.0344 & 0.430 $\pm$ 0.0166 &
    0.247 $\pm$ 0.0009 & 0.299 $\pm$ 0.0018 &
    0.170 $\pm$ 0.0000 & 0.269 $\pm$ 0.0001 &
    0.435 $\pm$ 0.0000 & 0.300 $\pm$ 0.0000 \\
 & FreqPool &
    0.286 $\pm$ 0.0127 & 0.377 $\pm$ 0.0081 &
    0.261 $\pm$ 0.0009 & 0.316 $\pm$ 0.0017 &
    0.182 $\pm$ 0.0000 & 0.280 $\pm$ 0.0002 &
    0.466 $\pm$ 0.0000 & 0.332 $\pm$ 0.0001 \\
 & Upsample &
    \secondres{0.242 $\pm$ 0.0082} & \secondres{0.351 $\pm$ 0.0067} &
    0.246 $\pm$ 0.0006 & 0.298 $\pm$ 0.0010 &
    0.217 $\pm$ 0.0004 & 0.310 $\pm$ 0.0005 &
    0.519 $\pm$ 0.0007 & 0.397 $\pm$ 0.0011 \\
 & STAug &
    0.383 $\pm$ 0.0449 & 0.421 $\pm$ 0.0183 &
    0.345 $\pm$ 0.0164 & 0.391 $\pm$ 0.0151 &
    - & - &
    - & - \\
 & Freq-Mask/Mix &
    0.317 $\pm$ 0.0146 & 0.393 $\pm$ 0.0066 &
    0.245 $\pm$ 0.0008 & \secondres{0.296 $\pm$ 0.0017} &
    0.167 $\pm$ 0.0000 & 0.265 $\pm$ 0.0002 &
    0.436 $\pm$ 0.0000 & 0.298 $\pm$ 0.0000 \\
 & Wave-Mask/Mix &
    0.375 $\pm$ 0.0336 & 0.418 $\pm$ 0.0142 &
    0.245 $\pm$ 0.0006 & 0.297 $\pm$ 0.0012 &
    \secondres{0.166 $\pm$ 0.0000} & 0.264 $\pm$ 0.0001 &
    \secondres{0.434 $\pm$ 0.0000} & 0.296 $\pm$ 0.0000 \\
 & Dominant Shuffle &
    0.340 $\pm$ 0.0525 & 0.408 $\pm$ 0.0220 &
    0.246 $\pm$ 0.0003 & 0.297 $\pm$ 0.0007 &
    0.167 $\pm$ 0.0000 & 0.265 $\pm$ 0.0002 &
    0.435 $\pm$ 0.0001 & 0.297 $\pm$ 0.0003 \\
  \rowcolor{tpsblue} &   \textbf{TPS (Ours)} &
    \boldres{0.237 $\pm$ 0.0035} & \boldres{0.349 $\pm$ 0.0026} &
    \boldres{0.239 $\pm$ 0.0003} & \boldres{0.285 $\pm$ 0.0007} &
    \boldres{0.166 $\pm$ 0.0000} & \boldres{0.263 $\pm$ 0.0002} &
    \boldres{0.432 $\pm$ 0.0000} & \boldres{0.292 $\pm$ 0.0000} \\
\midrule
\multirow{11}{*}{\rotatebox{90}{\textbf{PatchTST}}}
 & None &
    0.381 $\pm$ 0.0082 & 0.413 $\pm$ 0.0052 &
    0.237 $\pm$ 0.0008 & 0.272 $\pm$ 0.0007 &
    0.163 $\pm$ 0.0003 & 0.255 $\pm$ 0.0003 &
    0.411 $\pm$ 0.0005 & 0.274 $\pm$ 0.0005 \\
 & wDBA &
    0.391 $\pm$ 0.0165 & 0.418 $\pm$ 0.0078 &
    \secondres{0.230 $\pm$ 0.0005} & \boldres{0.266 $\pm$ 0.0003} &
    0.162 $\pm$ 0.0004 & 0.255 $\pm$ 0.0004 &
    0.412 $\pm$ 0.0005 & 0.274 $\pm$ 0.0004 \\
 & MBB &
    0.391 $\pm$ 0.0091 & 0.417 $\pm$ 0.0053 &
    0.231 $\pm$ 0.0006 & 0.266 $\pm$ 0.0009 &
    0.164 $\pm$ 0.0004 & 0.257 $\pm$ 0.0003 &
    0.413 $\pm$ 0.0004 & 0.275 $\pm$ 0.0005 \\
 & RobustTAD-m/p &
    0.395 $\pm$ 0.0152 & 0.419 $\pm$ 0.0074 &
    0.236 $\pm$ 0.0007 & 0.271 $\pm$ 0.0008 &
    \secondres{0.162 $\pm$ 0.0001} & \secondres{0.254 $\pm$ 0.0002} &
    0.408 $\pm$ 0.0004 & \secondres{0.273 $\pm$ 0.0004} \\
 & FreqAdd &
    0.375 $\pm$ 0.0083 & 0.414 $\pm$ 0.0051 &
    0.239 $\pm$ 0.0007 & 0.274 $\pm$ 0.0007 &
    0.164 $\pm$ 0.0003 & 0.258 $\pm$ 0.0003 &
    0.408 $\pm$ 0.0004 & 0.275 $\pm$ 0.0004 \\
 & FreqPool &
    0.401 $\pm$ 0.0136 & 0.424 $\pm$ 0.0063 &
    0.238 $\pm$ 0.0007 & 0.273 $\pm$ 0.0009 &
    0.163 $\pm$ 0.0005 & 0.255 $\pm$ 0.0003 &
    0.412 $\pm$ 0.0006 & 0.278 $\pm$ 0.0010 \\
 & Upsample &
    \secondres{0.361 $\pm$ 0.0064} & 0.415 $\pm$ 0.0039 &
    0.234 $\pm$ 0.0007 & 0.270 $\pm$ 0.0010 &
    0.165 $\pm$ 0.0006 & 0.257 $\pm$ 0.0006 &
    0.415 $\pm$ 0.0007 & 0.281 $\pm$ 0.0009 \\
 & STAug &
    0.381 $\pm$ 0.0086 & \secondres{0.413 $\pm$ 0.0045} &
    0.265 $\pm$ 0.0011 & 0.303 $\pm$ 0.0013 &
    - & - &
    - & - \\
 & Freq-Mask/Mix &
    0.410 $\pm$ 0.0083 & 0.428 $\pm$ 0.0050 &
    0.238 $\pm$ 0.0005 & 0.273 $\pm$ 0.0007 &
    0.164 $\pm$ 0.0001 & 0.256 $\pm$ 0.0002 &
    0.412 $\pm$ 0.0003 & 0.276 $\pm$ 0.0003 \\
 & Wave-Mask/Mix &
    0.380 $\pm$ 0.0103 & 0.426 $\pm$ 0.0065 &
    0.236 $\pm$ 0.0004 & 0.271 $\pm$ 0.0005 &
    0.163 $\pm$ 0.0002 & 0.256 $\pm$ 0.0002 &
    \boldres{0.407 $\pm$ 0.0002} & \boldres{0.273 $\pm$ 0.0002} \\
 & Dominant Shuffle &
    0.367 $\pm$ 0.0082 & 0.418 $\pm$ 0.0036 &
    0.241 $\pm$ 0.0006 & 0.275 $\pm$ 0.0051 &
    0.163 $\pm$ 0.0002 & 0.255 $\pm$ 0.0001 &
    0.409 $\pm$ 0.0004 & 0.274 $\pm$ 0.0002 \\
  \rowcolor{tpsblue} &  \textbf{TPS (Ours)} &
    \boldres{0.346 $\pm$ 0.0028} & \boldres{0.397 $\pm$ 0.0019} &
    \boldres{0.230 $\pm$ 0.0005} & \secondres{0.266 $\pm$ 0.0009} &
    \boldres{0.162 $\pm$ 0.0001} & \boldres{0.254 $\pm$ 0.0002} &
    \secondres{0.408 $\pm$ 0.0004} & 0.274 $\pm$ 0.0004 \\
\midrule
\multirow{11}{*}{\rotatebox{90}{\textbf{TiDE}}}
 & None &
    0.382 $\pm$ 0.0023 & 0.414 $\pm$ 0.0010 &
    0.240 $\pm$ 0.0004 & 0.279 $\pm$ 0.0005 &
    0.162 $\pm$ 0.0000 & \secondres{0.255 $\pm$ 0.0000} &
    0.440 $\pm$ 0.0003 & 0.319 $\pm$ 0.0003 \\
 & wDBA &
    0.379 $\pm$ 0.0011 & 0.412 $\pm$ 0.0010 &
    0.240 $\pm$ 0.0004 & 0.277 $\pm$ 0.0002 &
    0.162 $\pm$ 0.0000 & 0.255 $\pm$ 0.0002 &
    0.440 $\pm$ 0.0005 & 0.320 $\pm$ 0.0005 \\
 & MBB &
    0.380 $\pm$ 0.0029 & 0.413 $\pm$ 0.0008 &
    0.240 $\pm$ 0.0004 & 0.279 $\pm$ 0.0003 &
    0.165 $\pm$ 0.0003 & 0.259 $\pm$ 0.0004 &
    0.442 $\pm$ 0.0005 & 0.321 $\pm$ 0.0003 \\
 & RobustTAD-m/p &
    0.385 $\pm$ 0.0023 & 0.415 $\pm$ 0.0003 &
    0.240 $\pm$ 0.0004 & 0.279 $\pm$ 0.0005 &
    0.162 $\pm$ 0.0000 & 0.255 $\pm$ 0.0001 &
    0.440 $\pm$ 0.0005 & 0.319 $\pm$ 0.0004 \\
 & FreqAdd &
    0.381 $\pm$ 0.0015 & 0.415 $\pm$ 0.0010 &
    0.242 $\pm$ 0.0004 & 0.282 $\pm$ 0.0005 &
    0.165 $\pm$ 0.0000 & 0.259 $\pm$ 0.0000 &
    \boldres{0.436 $\pm$ 0.0003} & \boldres{0.314 $\pm$ 0.0004} \\
 & FreqPool &
    0.387 $\pm$ 0.0020 & 0.416 $\pm$ 0.0016 &
    0.253 $\pm$ 0.0006 & 0.290 $\pm$ 0.0002 &
    0.172 $\pm$ 0.0002 & 0.267 $\pm$ 0.0002 &
    0.465 $\pm$ 0.0004 & 0.343 $\pm$ 0.0005 \\
 & Upsample &
    \secondres{0.372 $\pm$ 0.0014} & \secondres{0.409 $\pm$ 0.0007} &
    \secondres{0.238 $\pm$ 0.0006} & \secondres{0.275 $\pm$ 0.0011} &
    0.181 $\pm$ 0.0002 & 0.277 $\pm$ 0.0004 &
    0.488 $\pm$ 0.0007 & 0.372 $\pm$ 0.0006 \\
 & STAug &
    0.383 $\pm$ 0.0017 & 0.414 $\pm$ 0.0013 &
    0.341 $\pm$ 0.0213 & 0.338 $\pm$ 0.0106 &
    - & - &
    - & - \\
 & Freq-Mask/Mix &
    0.421 $\pm$ 0.0071 & 0.430 $\pm$ 0.0023 &
    0.242 $\pm$ 0.0004 & 0.280 $\pm$ 0.0004 &
    0.164 $\pm$ 0.0000 & 0.258 $\pm$ 0.0000 &
    0.439 $\pm$ 0.0003 & 0.318 $\pm$ 0.0031 \\
 & Wave-Mask/Mix &
    0.382 $\pm$ 0.0029 & 0.413 $\pm$ 0.0011 &
    0.241 $\pm$ 0.0004 & 0.280 $\pm$ 0.0004 &
    \secondres{0.162 $\pm$ 0.0000} & 0.255 $\pm$ 0.0001 &
    0.438 $\pm$ 0.0003 & 0.318 $\pm$ 0.0003 \\
 & Dominant Shuffle &
    0.376 $\pm$ 0.0040 & 0.413 $\pm$ 0.0015 &
    0.243 $\pm$ 0.0003 & 0.281 $\pm$ 0.0003 &
    0.163 $\pm$ 0.0000 & 0.257 $\pm$ 0.0000 &
    0.439 $\pm$ 0.0003 & 0.318 $\pm$ 0.0003 \\
  \rowcolor{tpsblue} &  \textbf{TPS (Ours)} &
    \boldres{0.367 $\pm$ 0.0009} & \boldres{0.407 $\pm$ 0.0003} &
    \boldres{0.234 $\pm$ 0.0001} & \boldres{0.272 $\pm$ 0.0013} &
    \boldres{0.162 $\pm$ 0.0000} & \boldres{0.255 $\pm$ 0.0000} &
    \secondres{0.437 $\pm$ 0.0002} & \secondres{0.316 $\pm$ 0.0002} \\
\midrule
\multirow{11}{*}{\rotatebox{90}{\textbf{LightTS}}}
 & None &
    0.416 $\pm$ 0.0348 & 0.462 $\pm$ 0.0133 &
    0.236 $\pm$ 0.0027 & 0.290 $\pm$ 0.0036 &
    0.182 $\pm$ 0.0088 & 0.288 $\pm$ 0.0106 &
    0.445 $\pm$ 0.0053 & 0.328 $\pm$ 0.0054 \\
 & wDBA &
    0.414 $\pm$ 0.0214 & 0.453 $\pm$ 0.0110 &
    0.235 $\pm$ 0.0031 & 0.287 $\pm$ 0.0030 &
    0.183 $\pm$ 0.0077 & 0.290 $\pm$ 0.0088 &
    0.445 $\pm$ 0.0042 & 0.327 $\pm$ 0.0034 \\
 & MBB &
    0.427 $\pm$ 0.0182 & 0.462 $\pm$ 0.0102 &
    0.237 $\pm$ 0.0026 & 0.290 $\pm$ 0.0031 &
    0.185 $\pm$ 0.0057 & 0.294 $\pm$ 0.0434 &
    0.449 $\pm$ 0.0058 & 0.330 $\pm$ 0.0030 \\
 & RobustTAD-m/p &
    0.419 $\pm$ 0.0304 & 0.459 $\pm$ 0.0156 &
    0.236 $\pm$ 0.0031 & 0.290 $\pm$ 0.0033 &
    0.180 $\pm$ 0.0064 & 0.285 $\pm$ 0.0082 &
    0.440 $\pm$ 0.0104 & \boldres{0.319 $\pm$ 0.0105} \\
 & FreqAdd &
    0.425 $\pm$ 0.0374 & 0.465 $\pm$ 0.0196 &
    0.238 $\pm$ 0.0028 & 0.293 $\pm$ 0.0046 &
    0.179 $\pm$ 0.0065 & 0.284 $\pm$ 0.0070 &
    0.448 $\pm$ 0.0068 & 0.329 $\pm$ 0.0067 \\
 & FreqPool &
    0.326 $\pm$ 0.0977 & 0.396 $\pm$ 0.0075 &
    0.239 $\pm$ 0.0019 & 0.288 $\pm$ 0.0015 &
    0.186 $\pm$ 0.0072 & 0.290 $\pm$ 0.0075 &
    0.468 $\pm$ 0.0087 & 0.347 $\pm$ 0.0060 \\
 & Upsample &
    \secondres{0.289 $\pm$ 0.0128} & \secondres{0.394 $\pm$ 0.0080} &
    0.235 $\pm$ 0.0028 & \secondres{0.285 $\pm$ 0.0018} &
    0.187 $\pm$ 0.0058 & 0.294 $\pm$ 0.0072 &
    0.474 $\pm$ 0.0102 & 0.354 $\pm$ 0.0088 \\
 & STAug &
    0.410 $\pm$ 0.0366 & 0.452 $\pm$ 0.0160 &
    0.287 $\pm$ 0.0144 & 0.339 $\pm$ 0.0165 &
    - & - &
    - & - \\
 & Freq-Mask/Mix &
    0.372 $\pm$ 0.0360 & 0.440 $\pm$ 0.0096 &
    \secondres{0.235 $\pm$ 0.0020} & 0.287 $\pm$ 0.0026 &
    0.179 $\pm$ 0.0062 & \secondres{0.282 $\pm$ 0.0061} &
    0.441 $\pm$ 0.0027 & 0.322 $\pm$ 0.0023 \\
 & Wave-Mask/Mix &
    0.421 $\pm$ 0.0310 & 0.453 $\pm$ 0.0142 &
    0.235 $\pm$ 0.0039 & 0.289 $\pm$ 0.0059 &
    \secondres{0.178 $\pm$ 0.0059} & 0.284 $\pm$ 0.0077 &
    \secondres{0.440 $\pm$ 0.0047} & 0.322 $\pm$ 0.0039 \\
 & Dominant Shuffle &
    0.358 $\pm$ 0.0138 & 0.432 $\pm$ 0.0089 &
    0.235 $\pm$ 0.0032 & 0.287 $\pm$ 0.0034 &
    0.192 $\pm$ 0.0111 & 0.296 $\pm$ 0.0093 &
    0.445 $\pm$ 0.0051 & 0.327 $\pm$ 0.0055 \\
 \rowcolor{tpsblue} & \textbf{TPS (Ours)} &
    \boldres{0.273 $\pm$ 0.0183} & \boldres{0.385 $\pm$ 0.0118} &
    \boldres{0.228 $\pm$ 0.0026} & \boldres{0.277 $\pm$ 0.0019} &
    \boldres{0.174 $\pm$ 0.0030} & \boldres{0.278 $\pm$ 0.0050} &
    \boldres{0.439 $\pm$ 0.0071} & \secondres{0.320 $\pm$ 0.0055} \\
\bottomrule
\end{tabular}
}
\end{table*}

\subsection{Short-Term Forecasting Per Prediction Length} \label{appendix:C2}

Table~\ref{tab:appendix_short} reports the short-term forecasting results (MSE and MAE) on PeMS-\{03, 04, 07, 08\} for prediction lengths \{12, 24, 36, 48\}, averaged over five runs. TPS achieves the best performance in most cases, except at the 48-step prediction length where it generally ranks second. The strongest competing methods are typically STAug, FreqPool, or Wave-Mask/Mix.

Experiments in this setting are conducted using PatchTST, which serves as a strong representative backbone on the PeMS benchmarks. This choice keeps the short-term evaluation tractable while still providing a meaningful comparison across augmentation methods on large, high-dimensional traffic datasets.

\begin{table*}[ht]
\centering
\scriptsize
\setlength{\tabcolsep}{4pt}
\caption{
Short-term traffic forecasting with PatchTST on PeMS-\{03,04,07,08\}.
For each prediction length, results are reported as mean$\pm$std over 5 runs.
Best results are highlighted in bold, and second-best results are underlined.
}
\label{tab:appendix_short}
\resizebox{\textwidth}{!}{
\begin{tabular}{c|c|cc|cc|cc|cc}
\toprule
\multirow{14}{*}{\rotatebox{90}{\textbf{PeMS03}}}
 & \textbf{Method}
     & \multicolumn{2}{c|}{12}
     & \multicolumn{2}{c|}{24}
     & \multicolumn{2}{c|}{36}
     & \multicolumn{2}{c}{48} \\
 &  & MSE & MAE & MSE & MAE & MSE & MAE & MSE & MAE \\
\midrule
 & None & 0.084 $\pm$ 0.0024 & 0.200 $\pm$ 0.0046
        & 0.115 $\pm$ 0.0054 & 0.238 $\pm$ 0.0115
        & 0.127 $\pm$ 0.0037 & 0.239 $\pm$ 0.0038
        & 0.147 $\pm$ 0.0072 & 0.259 $\pm$ 0.0101 \\
 & wDBA~\cite{asd} & 0.086 $\pm$ 0.0060 & 0.203 $\pm$ 0.0092
       & 0.120 $\pm$ 0.0043 & 0.230 $\pm$ 0.0014
       & 0.135 $\pm$ 0.0070 & 0.265 $\pm$ 0.0106
       & 0.158 $\pm$ 0.0013 & 0.289 $\pm$ 0.0024 \\
 & MBB~\cite{mbb} &  \boldres{0.075 $\pm$ 0.0005} &  \secondres{0.183 $\pm$ 0.0008}
       & 0.107 $\pm$ 0.0017 & 0.221 $\pm$ 0.0025
       & 0.140 $\pm$ 0.0013 & 0.259 $\pm$ 0.0051
       & 0.151 $\pm$ 0.0013 & 0.262 $\pm$ 0.0015 \\
 & RobustTAD-m/p~\cite{gao2021robusttadrobusttimeseries} & 0.090 $\pm$ 0.0016 & 0.206 $\pm$ 0.0041
               & 0.118 $\pm$ 0.0041 & 0.240 $\pm$ 0.0062
               & 0.128 $\pm$ 0.0065 & 0.247 $\pm$ 0.0089
               & 0.145 $\pm$ 0.0036 & 0.257 $\pm$ 0.0020 \\
 & FreqAdd~\cite{freqadd} & 0.088 $\pm$ 0.0041 & 0.205 $\pm$ 0.0082
           & 0.119 $\pm$ 0.0063 & 0.244 $\pm$ 0.0110
           & 0.135 $\pm$ 0.0032 & 0.253 $\pm$ 0.0047
           & 0.149 $\pm$ 0.0047 & 0.261 $\pm$ 0.0036 \\
 & FreqPool~\cite{freqpool} & 0.078 $\pm$ 0.0009 & 0.190 $\pm$ 0.0022
            & 0.106 $\pm$ 0.0053 & 0.227 $\pm$ 0.0103
            & 0.120 $\pm$ 0.0027 & 0.237 $\pm$ 0.0036
            & 0.145 $\pm$ 0.0032 & 0.265 $\pm$ 0.0067 \\
 & Upsample~\cite{upsample} & 0.080 $\pm$ 0.0024 & 0.195 $\pm$ 0.0051
            & 0.110 $\pm$ 0.0066 & 0.235 $\pm$ 0.0115
            &  \secondres{0.119 $\pm$ 0.0054} & 0.236 $\pm$ 0.0069
            & 0.140 $\pm$ 0.0061 & 0.256 $\pm$ 0.0079 \\
 & STAug~\cite{zhang2023diversecoherentaugmentationtimeseries} &  \secondres{0.076 $\pm$ 0.0014} & 0.185 $\pm$ 0.0032
         &  \secondres{0.102 $\pm$ 0.0015} & \secondres{0.214 $\pm$ 0.0017}
         & 0.125 $\pm$ 0.0027 & 0.238 $\pm$ 0.0038
         & 0.147 $\pm$ 0.0029 & 0.260 $\pm$ 0.0051 \\
 & Freq-Mask/Mix~\cite{chen2023fraugfrequencydomainaugmentation} & 0.084 $\pm$ 0.0030 & 0.198 $\pm$ 0.0058
                & 0.123 $\pm$ 0.0067 & 0.253 $\pm$ 0.0142
                & 0.134 $\pm$ 0.0077 & 0.251 $\pm$ 0.0090
                & 0.153 $\pm$ 0.0082 & 0.268 $\pm$ 0.0114 \\
 & Wave-Mask/Mix~\cite{arabi2024wavemaskmixexploringwaveletbasedaugmentations} & 0.085 $\pm$ 0.0080 & 0.196 $\pm$ 0.0079
                & 0.106 $\pm$ 0.0029 & 0.221 $\pm$ 0.0044
                & 0.122 $\pm$ 0.0064 &  \secondres{0.235 $\pm$ 0.0064}
                & 0.145 $\pm$ 0.0061 & 0.258 $\pm$ 0.0070 \\
 & Dominant Shuffle~\cite{zhao2024dominantshufflesimplepowerful} & 0.087 $\pm$ 0.0009 & 0.199 $\pm$ 0.0039
               & 0.111 $\pm$ 0.0034 & 0.235 $\pm$ 0.0062
               & 0.122 $\pm$ 0.0021 & 0.238 $\pm$ 0.0032
               &  \secondres{0.139 $\pm$ 0.0044} &  \secondres{0.253 $\pm$ 0.0061} \\
  \rowcolor{tpsblue} &   \textbf{TPS (Ours)} & 0.076 $\pm$ 0.0024 & \boldres{0.181 $\pm$ 0.0019}
              & \boldres{0.094 $\pm$ 0.0017} & \boldres{0.207 $\pm$ 0.0029}
              & \boldres{0.114 $\pm$ 0.0023} & \boldres{0.230 $\pm$ 0.0039}
              & \boldres{0.133 $\pm$ 0.0056} & \boldres{0.245 $\pm$ 0.0105} \\
\midrule
\multirow{14}{*}{\rotatebox{90}{\textbf{PeMS04}}}
 & None
   & 0.094 $\pm$ 0.0007 & 0.207 $\pm$ 0.0008
   & 0.121 $\pm$ 0.0013 & 0.238 $\pm$ 0.0021
   & 0.155 $\pm$ 0.0070 & 0.272 $\pm$ 0.0101
   & 0.171 $\pm$ 0.0089 & 0.280 $\pm$ 0.0067 \\
 & wDBA
   & 0.093 $\pm$ 0.0019 & 0.205 $\pm$ 0.0044
   & 0.125 $\pm$ 0.0015 & 0.241 $\pm$ 0.0022
   & 0.150 $\pm$ 0.0033 & 0.259 $\pm$ 0.0036
   & 0.169 $\pm$ 0.0018 & 0.277 $\pm$ 0.0049 \\
 & MBB
   & 0.095 $\pm$ 0.0001 & 0.210 $\pm$ 0.0014
   & 0.133 $\pm$ 0.0081 & 0.253 $\pm$ 0.0116
   & 0.153 $\pm$ 0.0057 & 0.271 $\pm$ 0.0091
   & 0.176 $\pm$ 0.0045 & 0.285 $\pm$ 0.0056 \\
 & RobustTAD-m/p
   & 0.095 $\pm$ 0.0017 & 0.209 $\pm$ 0.0027
   & 0.125 $\pm$ 0.0054 & 0.241 $\pm$ 0.0068
   & 0.147 $\pm$ 0.0017 & 0.263 $\pm$ 0.0047
   & 0.168 $\pm$ 0.0030 & 0.281 $\pm$ 0.0047 \\
 & FreqAdd
   & 0.098 $\pm$ 0.0047 & 0.213 $\pm$ 0.0068
   & 0.128 $\pm$ 0.0036 & 0.244 $\pm$ 0.0037
   & 0.155 $\pm$ 0.0034 & 0.268 $\pm$ 0.0072
   & 0.175 $\pm$ 0.0042 & 0.284 $\pm$ 0.0062 \\
 & FreqPool
   & 0.093 $\pm$ 0.0015 &  \secondres{0.204 $\pm$ 0.0013}
   &  \secondres{0.119 $\pm$ 0.0014} & \secondres{0.235 $\pm$ 0.0032}
   & \secondres{0.142 $\pm$ 0.0037} & 0.260 $\pm$ 0.0056
   & \boldres{0.157 $\pm$ 0.0036} & \boldres{0.271 $\pm$ 0.0031} \\
 & Upsample
   & 0.096 $\pm$ 0.0020 & 0.210 $\pm$ 0.0044
   & 0.127 $\pm$ 0.0013 & 0.247 $\pm$ 0.0028
   & 0.149 $\pm$ 0.0039 & 0.272 $\pm$ 0.0083
   & 0.171 $\pm$ 0.0065 & 0.285 $\pm$ 0.0055 \\
 & STAug
   & 0.093 $\pm$ 0.0012 & 0.204 $\pm$ 0.0024
   & 0.122 $\pm$ 0.0026 & 0.238 $\pm$ 0.0044
   & 0.148 $\pm$ 0.0031 & 0.262 $\pm$ 0.0050
   & 0.178 $\pm$ 0.0043 & 0.290 $\pm$ 0.0036 \\
 & Freq-Mask/Mix
   & 0.101 $\pm$ 0.0005 & 0.218 $\pm$ 0.0018
   & 0.133 $\pm$ 0.0042 & 0.246 $\pm$ 0.0043
   & 0.158 $\pm$ 0.0071 & 0.272 $\pm$ 0.0070
   & 0.178 $\pm$ 0.0084 & 0.292 $\pm$ 0.0077 \\
 &  Wave-Mask/Mix
   &  \secondres{0.092 $\pm$ 0.0008} & 0.205 $\pm$ 0.0014
   & 0.119 $\pm$ 0.0036 & 0.235 $\pm$ 0.0038
   & 0.147 $\pm$ 0.0023 & \secondres{0.259 $\pm$ 0.0026}
   & 0.172 $\pm$ 0.0048 & 0.285 $\pm$ 0.0051 \\
 & Dominant Shuffle
   & 0.097 $\pm$ 0.0015 & 0.213 $\pm$ 0.0034
   & 0.123 $\pm$ 0.0026 & 0.240 $\pm$ 0.0017
   & 0.148 $\pm$ 0.0056 & 0.267 $\pm$ 0.0066
   & 0.169 $\pm$ 0.0018 & 0.287 $\pm$ 0.0041 \\
  \rowcolor{tpsblue} &   \textbf{TPS (Ours)}
   & \boldres{0.091 $\pm$ 0.0012} & \boldres{0.202 $\pm$ 0.0018}
   & \boldres{0.112 $\pm$ 0.0023} & \boldres{0.227 $\pm$ 0.0029}
   & \boldres{0.133 $\pm$ 0.0039} & \boldres{0.250 $\pm$ 0.0046}
   & \secondres{0.163 $\pm$ 0.0063} & \secondres{0.274 $\pm$ 0.0068} \\
\midrule
\multirow{14}{*}{\rotatebox{90}{\textbf{PeMS07}}}
 & None
   & 0.073 $\pm$ 0.0011 & 0.188 $\pm$ 0.0034
   & 0.101 $\pm$ 0.0061 & 0.222 $\pm$ 0.0100
   & 0.129 $\pm$ 0.0050 & 0.252 $\pm$ 0.0115
   & 0.164 $\pm$ 0.0059 & 0.300 $\pm$ 0.0081 \\
 & wDBA
   & 0.075 $\pm$ 0.0022 & 0.190 $\pm$ 0.0042
   & 0.100 $\pm$ 0.0007 & 0.224 $\pm$ 0.0053
   & 0.120 $\pm$ 0.0120 & 0.245 $\pm$ 0.0168
   & 0.142 $\pm$ 0.0050 & 0.273 $\pm$ 0.0038 \\
 & MBB
   & 0.078 $\pm$ 0.0010 & 0.201 $\pm$ 0.0030
   & 0.107 $\pm$ 0.0010 & 0.236 $\pm$ 0.0022
   & 0.143 $\pm$ 0.0062 & 0.284 $\pm$ 0.0051
   & 0.172 $\pm$ 0.0042 & 0.310 $\pm$ 0.0048 \\
 &  RobustTAD-m/p
   & 0.076 $\pm$ 0.0020 & 0.190 $\pm$ 0.0044
   & 0.103 $\pm$ 0.0036 & 0.231 $\pm$ 0.0076
   & 0.120 $\pm$ 0.0027 & 0.246 $\pm$ 0.0043
   & 0.146 $\pm$ 0.0069 & 0.274 $\pm$ 0.0117 \\
 & FreqAdd
   & 0.076 $\pm$ 0.0018 & 0.194 $\pm$ 0.0044
   & 0.106 $\pm$ 0.0033 & 0.238 $\pm$ 0.0078
   & 0.136 $\pm$ 0.0100 & 0.273 $\pm$ 0.0153
   & 0.154 $\pm$ 0.0118 & 0.290 $\pm$ 0.0179 \\
 & FreqPool
   & 0.076 $\pm$ 0.0095 & 0.187 $\pm$ 0.0116
   & \secondres{0.095 $\pm$ 0.0040} & 0.219 $\pm$ 0.0093
   & 0.124 $\pm$ 0.0164 & 0.253 $\pm$ 0.0273
   & 0.137 $\pm$ 0.0099 & 0.268 $\pm$ 0.0161 \\
 & Upsample
   & 0.078 $\pm$ 0.0060 & 0.193 $\pm$ 0.0100
   & 0.104 $\pm$ 0.0043 & 0.228 $\pm$ 0.0079
   & 0.121 $\pm$ 0.0076 & 0.248 $\pm$ 0.0124
   & \secondres{0.134 $\pm$ 0.0063} & 0.263 $\pm$ 0.0099 \\
 & STAug
   & \secondres{0.071 $\pm$ 0.0020} & \secondres{0.181 $\pm$ 0.0060}
   & 0.114 $\pm$ 0.0054 & \secondres{0.218 $\pm$ 0.0045}
   & 0.129 $\pm$ 0.0082 & 0.255 $\pm$ 0.0075
   & 0.159 $\pm$ 0.0054 & 0.285 $\pm$ 0.0076 \\
 & Freq-Mask/Mix
   & 0.079 $\pm$ 0.0037 & 0.194 $\pm$ 0.0060
   & 0.110 $\pm$ 0.0074 & 0.239 $\pm$ 0.0130
   & 0.126 $\pm$ 0.0047 & 0.249 $\pm$ 0.0051
   & 0.137 $\pm$ 0.0075 & \secondres{0.258 $\pm$ 0.0093} \\
 &  Wave-Mask/Mix
   & 0.074 $\pm$ 0.0053 & 0.190 $\pm$ 0.0128
   & 0.096 $\pm$ 0.0025 & 0.221 $\pm$ 0.0040
   & \secondres{0.118 $\pm$ 0.0037} & 0.248 $\pm$ 0.0068
   & \boldres{0.130 $\pm$ 0.0022} & \boldres{0.254 $\pm$ 0.0032} \\
 & Dominant Shuffle
   & 0.078 $\pm$ 0.0031 & 0.195 $\pm$ 0.0080
   & 0.100 $\pm$ 0.0063 & 0.222 $\pm$ 0.0113
   & 0.118 $\pm$ 0.0076 & \secondres{0.244 $\pm$ 0.0124}
   & 0.139 $\pm$ 0.0095 & 0.264 $\pm$ 0.0156 \\
  \rowcolor{tpsblue} &   \textbf{TPS (Ours)}
   & \boldres{0.070 $\pm$ 0.0023} & \boldres{0.179 $\pm$ 0.0038}
   & \boldres{0.090 $\pm$ 0.0030} & \boldres{0.207 $\pm$ 0.0045}
   & \boldres{0.118 $\pm$ 0.0113} & \boldres{0.244 $\pm$ 0.0203}
   & 0.143 $\pm$ 0.0075 & 0.270 $\pm$ 0.0120 \\
\midrule
\multirow{14}{*}{\rotatebox{90}{\textbf{PeMS08}}}
 & None
   & 0.099 $\pm$ 0.0056 & 0.206 $\pm$ 0.0080
   & 0.143 $\pm$ 0.0034 & 0.243 $\pm$ 0.0061
   & 0.186 $\pm$ 0.0050 & 0.285 $\pm$ 0.0117
   & 0.206 $\pm$ 0.0140 & 0.302 $\pm$ 0.0123 \\
 & wDBA
   & 0.102 $\pm$ 0.0069 & 0.210 $\pm$ 0.0128
   & 0.146 $\pm$ 0.0045 & 0.239 $\pm$ 0.0053
   & 0.200 $\pm$ 0.0035 & 0.282 $\pm$ 0.0198
   & 0.232 $\pm$ 0.0124 & 0.316 $\pm$ 0.0175 \\
 & MBB
   & 0.095 $\pm$ 0.0006 & 0.203 $\pm$ 0.0012
   & 0.149 $\pm$ 0.0044 & 0.245 $\pm$ 0.0092
   & 0.188 $\pm$ 0.0031 & 0.269 $\pm$ 0.0015
   & 0.244 $\pm$ 0.0074 & 0.324 $\pm$ 0.0161 \\
 & RobustTAD-m/p
   & 0.101 $\pm$ 0.0088 & 0.210 $\pm$ 0.0099
   & 0.145 $\pm$ 0.0036 & 0.242 $\pm$ 0.0035
   & 0.182 $\pm$ 0.0052 & 0.276 $\pm$ 0.0124
   & 0.208 $\pm$ 0.0155 & 0.308 $\pm$ 0.0128 \\
 & FreqAdd
   & 0.103 $\pm$ 0.0058 & 0.217 $\pm$ 0.0082
   & 0.151 $\pm$ 0.0069 & 0.255 $\pm$ 0.0089
   & 0.185 $\pm$ 0.0072 & 0.296 $\pm$ 0.0076
   & 0.206 $\pm$ 0.0100 & 0.304 $\pm$ 0.0108 \\
 & FreqPool
   & 0.097 $\pm$ 0.0031 & 0.203 $\pm$ 0.0044
   & \secondres{0.128 $\pm$ 0.0093} & 0.238 $\pm$ 0.0096
   & 0.161 $\pm$ 0.0028 & 0.270 $\pm$ 0.0130
   & \secondres{0.187 $\pm$ 0.0076} & 0.296 $\pm$ 0.0096 \\
 & Upsample
   & 0.093 $\pm$ 0.0039 & 0.205 $\pm$ 0.0053
   & 0.129 $\pm$ 0.0036 & 0.240 $\pm$ 0.0034
   & \secondres{0.160 $\pm$ 0.0110} & \secondres{0.267 $\pm$ 0.0119}
   & \boldres{0.181 $\pm$ 0.0084} & \boldres{0.279 $\pm$ 0.0093} \\
 & STAug
   & 0.094 $\pm$ 0.0050 & 0.205 $\pm$ 0.0089
   & 0.131 $\pm$ 0.0042 & \secondres{0.236 $\pm$ 0.0051}
   & 0.167 $\pm$ 0.0070 & 0.268 $\pm$ 0.0114
   & 0.206 $\pm$ 0.0077 & \secondres{0.295 $\pm$ 0.0093} \\
 & Freq-Mask/Mix
   & 0.101 $\pm$ 0.0014 & 0.206 $\pm$ 0.0022
   & 0.150 $\pm$ 0.0051 & 0.252 $\pm$ 0.0050
   & 0.194 $\pm$ 0.0102 & 0.302 $\pm$ 0.0139
   & 0.221 $\pm$ 0.0112 & 0.304 $\pm$ 0.0187 \\
 & Wave-Mask/Mix
   & \secondres{0.092 $\pm$ 0.0024} & \secondres{0.202 $\pm$ 0.0046}
   & 0.130 $\pm$ 0.0077 & 0.248 $\pm$ 0.0089
   & 0.167 $\pm$ 0.0149 & 0.276 $\pm$ 0.0156
   & 0.206 $\pm$ 0.0272 & 0.306 $\pm$ 0.0310 \\
 & Dominant Shuffle
   & 0.096 $\pm$ 0.0031 & 0.205 $\pm$ 0.0041
   & 0.141 $\pm$ 0.0072 & 0.244 $\pm$ 0.0074
   & 0.178 $\pm$ 0.0088 & 0.280 $\pm$ 0.0091
   & 0.210 $\pm$ 0.0043 & 0.293 $\pm$ 0.0052 \\
  \rowcolor{tpsblue} &   \textbf{TPS (Ours)}
   & \boldres{0.089 $\pm$ 0.0022} & \boldres{0.197 $\pm$ 0.0036}
   & \boldres{0.113 $\pm$ 0.0033} & \boldres{0.223 $\pm$ 0.0073}
   & \boldres{0.139 $\pm$ 0.0047} & \boldres{0.246 $\pm$ 0.0050}
   & 0.198 $\pm$ 0.0104 & 0.295 $\pm$ 0.0097 \\
\bottomrule
\end{tabular}
}
\end{table*}

\section{Further Ablation Studies} \label{appendix:ablation}

\paragraph{Full Component-wise Analysis.}
Table~\ref{tab:additional_ablation_dlinear} presents the complete component-wise ablation results on the ETT datasets, including MSE and MAE with their corresponding standard deviations. For each prediction length in \{96, 192, 336, 720\}, results are computed over five runs and then averaged across the four prediction lengths. The MAE results follow the same trends as the MSE results, further supporting the consistency of each component's contribution.

First, we remove the variance-based ordering used to prioritize patches for shuffling. The results indicate that this component generally provides a modest improvement, although its effect naturally vanishes when the shuffle rate is set to $1.0$.

Second, we replace overlapping patches with non-overlapping ones, which leads to a substantial degradation in performance. This finding supports our design choice in TPS: overlapping patches are important for preserving local temporal structure and reducing discontinuities at patch boundaries.

Third, we examine the role of data--label coherence in the augmentation pipeline. In the default design, augmentation is applied jointly to the input and forecast horizon. When augmentation is applied only to the input, performance deteriorates significantly, consistent with prior observations~\cite{chen2023fraugfrequencydomainaugmentation, wei2020circumventingoutliersautoaugmentknowledge} on the importance of maintaining input--target alignment.

Finally, we evaluate a frequency-domain variant in which the same patch-based operations are applied after transforming the signal using the Fast Fourier Transform (FFT). This variant also degrades performance, indicating that TPS is more effective in the time domain.

\begin{table*}[ht]
\caption{
Component-wise ablation results of TPS on ETT-\{h1,h2,m1,m2\} using DLinear.
Values are reported as mean $\pm$ std MSE/MAE over five runs, averaged across prediction lengths \{96, 192, 336, 720\}.
Best results are highlighted in bold, and second-best results are underlined.
}
\label{tab:additional_ablation_dlinear}
\centering
\scriptsize
\setlength{\tabcolsep}{3pt}
\renewcommand{\arraystretch}{1.2}
\begin{tabular}{l|cc|cc|cc|cc}
\toprule
\textbf{Methods}
& \multicolumn{2}{c|}{\textbf{ETTh1}}
& \multicolumn{2}{c|}{\textbf{ETTh2}}
& \multicolumn{2}{c|}{\textbf{ETTm1}}
& \multicolumn{2}{c}{\textbf{ETTm2}} \\
& MSE & MAE & MSE & MAE & MSE & MAE & MSE & MAE \\
\midrule
None
& 0.438 $\pm$ 0.0194 & 0.449 $\pm$ 0.0162
& 0.464 $\pm$ 0.0099 & 0.462 $\pm$ 0.0053
& 0.361 $\pm$ 0.0010 & \secondres{0.383 $\pm$ 0.0015}
& 0.276 $\pm$ 0.0099 & 0.339 $\pm$ 0.0093 \\
 \rowcolor{tpsblue} \textbf{TPS}
& \boldres{0.410 $\pm$ 0.0036} & \boldres{0.425 $\pm$ 0.0031}
& \boldres{0.369 $\pm$ 0.0056} & \boldres{0.408 $\pm$ 0.0039}
& \boldres{0.354 $\pm$ 0.0006} & \boldres{0.377 $\pm$ 0.0006}
& \boldres{0.261 $\pm$ 0.0018} & \boldres{0.324 $\pm$ 0.0027} \\
\hspace{0.5em}- Variance Score
& 0.417 $\pm$ 0.0166 & 0.430 $\pm$ 0.0129
& \secondres{0.370 $\pm$ 0.0066} & \secondres{0.409 $\pm$ 0.0048}
& \secondres{0.355 $\pm$ 0.0006} & \boldres{0.377 $\pm$ 0.0005}
& \boldres{0.261 $\pm$ 0.0024} & \boldres{0.324 $\pm$ 0.0033} \\
\hspace{0.5em}- Temporal Patching
& \secondres{0.416 $\pm$ 0.0083} & \secondres{0.430 $\pm$ 0.0040}
& 0.379 $\pm$ 0.0092 & 0.424 $\pm$ 0.0057
& 0.376 $\pm$ 0.0018 & 0.397 $\pm$ 0.0009
& \secondres{0.267 $\pm$ 0.0031} & \secondres{0.332 $\pm$ 0.0039} \\
\hspace{0.5em}- Data--Label Coherence
& 0.443 $\pm$ 0.0158 & 0.451 $\pm$ 0.0144
& 0.438 $\pm$ 0.0146 & 0.447 $\pm$ 0.0074
& 0.364 $\pm$ 0.0018 & 0.386 $\pm$ 0.0022
& 0.290 $\pm$ 0.0122 & 0.352 $\pm$ 0.0103 \\
\hspace{0.5em}+ Frequency Domain
& 0.437 $\pm$ 0.0096 & 0.448 $\pm$ 0.0077
& 0.470 $\pm$ 0.0267 & 0.464 $\pm$ 0.0132
& 0.363 $\pm$ 0.0011 & 0.384 $\pm$ 0.0016
& 0.285 $\pm$ 0.0082 & 0.345 $\pm$ 0.0065 \\
\bottomrule
\end{tabular}
\end{table*}

\begin{figure*}[ht]
\begin{center}
    \centerline{\includegraphics[width=\textwidth]{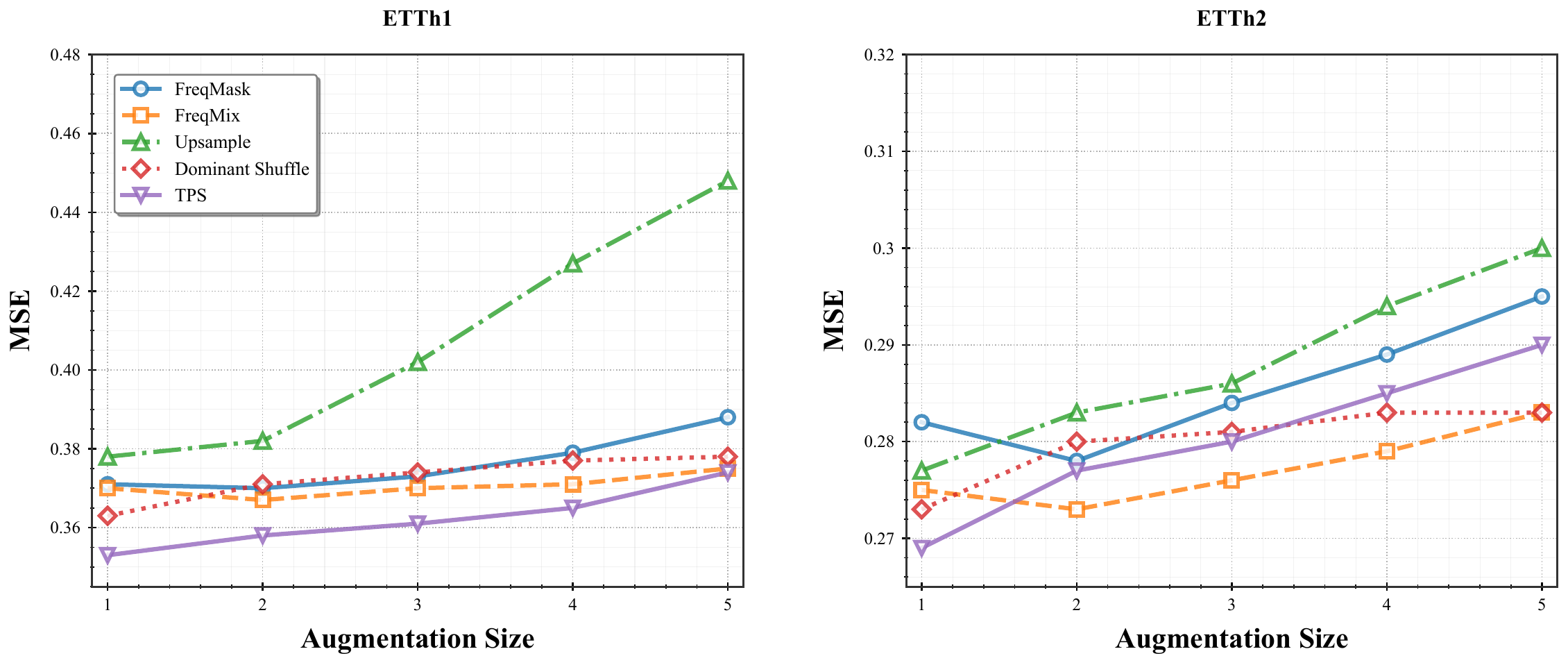}}
\caption{Impact of varying augmentation sizes (1--5) on forecasting performance using the PatchTST model with a prediction length of 96 on the ETTh1 and ETTh2 datasets. The results demonstrate how different augmentation methods respond to increasing augmentation intensity, highlighting stability or degradation in performance.}
\Description{
Two line plots comparing forecasting MSE on ETTh1 and ETTh2 as
augmentation size increases from one to five for FreqMask, FreqMix,
Upsample, Dominant Shuffle, and TPS.
}
    \label{fig:augsizetsf}
\end{center}
\end{figure*}

\paragraph{Impact of Varying Augmentation Sizes.}
Figure~\ref{fig:augsizetsf} reports an ablation study with varying augmentation sizes (1, 2, 3, 4, and 5) using PatchTST on ETTh1 and ETTh2 with prediction length 96. An augmentation size of 2 means that the augmented sample set is doubled by applying the augmentation method twice. This analysis examines whether increasing augmentation intensity continues to improve performance or instead introduces overly strong perturbations.

The results show that FreqMix benefits from larger augmentation sizes on both datasets, while FreqMask improves only on ETTh2 when applied twice. In contrast, most other methods degrade as augmentation size increases. Notably, TPS on ETTh1 exhibits only minor performance variation up to augmentation size 4, indicating stable behavior under stronger augmentation. This contrasts with methods such as Upsample and FreqMask, which are more sensitive to augmentation size. Dominant Shuffle and FreqMix also remain relatively stable across different augmentation sizes. On ETTh2, TPS shows some degradation at larger augmentation sizes, suggesting that overly strong perturbations can become harmful in this setting. Nevertheless, across other models and prediction lengths on ETTh2, TPS is generally stable under different augmentation sizes.

\begin{figure*}[ht]
\begin{center}
    \centerline{\includegraphics[width=\textwidth]{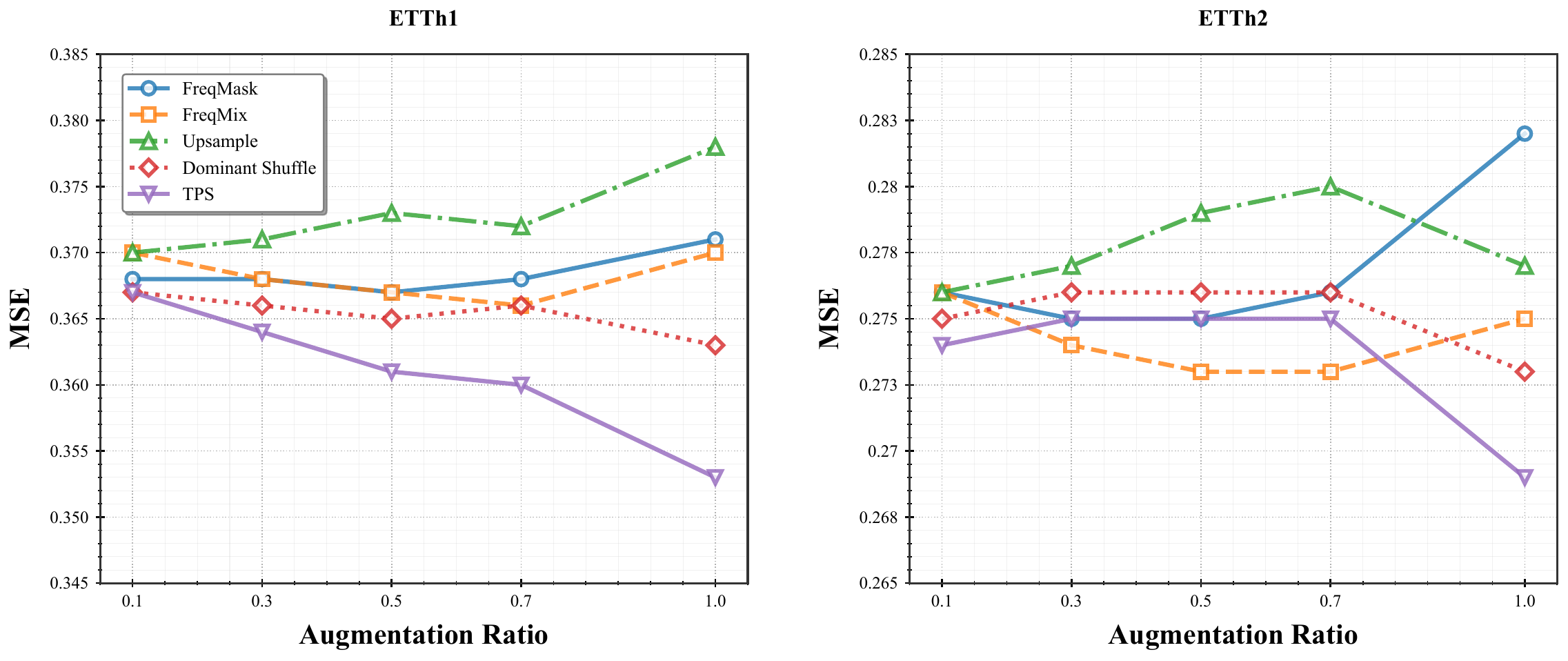}}
\caption{Effect of varying augmentation ratios (0.1 to 1.0) on the MSE performance of different augmentation methods using the PatchTST model with prediction length 96 on the ETTh1 and ETTh2 datasets. The augmentation ratio indicates the proportion of augmented samples used during training.}
\Description{
Two line plots comparing forecasting MSE on ETTh1 and ETTh2 across
augmentation ratios from 0.1 to 1.0 for FreqMask, FreqMix, Upsample,
Dominant Shuffle, and TPS.
}
    \label{fig:augratio}
\end{center}
\end{figure*}

\paragraph{Impact of Varying Augmentation Ratios.}
We also conduct experiments with different augmentation ratios (0.1, 0.3, 0.5, 0.7, and 1.0) using PatchTST on ETTh1 and ETTh2 with prediction length 96, as shown in Figure~\ref{fig:augratio}. The evaluated methods include FreqMask, FreqMix, Upsample, Dominant Shuffle, and TPS. Here, the augmentation ratio denotes the proportion of augmented samples included in each training batch, so lower ratios correspond to fewer augmented samples. TPS achieves the lowest MSE on both datasets when using the full augmentation ratio (1.0). Moreover, even with only 10\% augmented samples (ratio 0.1), TPS outperforms all competing augmentation methods at their respective ratios.

\section{TPS for Time Series Classification} \label{appendix:tpsclass}

\paragraph{Datasets and Experimental Setups.}
We use the widely adopted UCR~\cite{UCRArchive2018} and UEA~\cite{bagnall2018ueamultivariatetimeseries} repositories to evaluate TPS on univariate and multivariate time series classification tasks.

For univariate classification, we use the UCR archive, which contains datasets with a single time-dependent variable (i.e., one channel). The archive covers diverse categories, including Device, ECG, Image, Motion, Sensor, Spectrograph, Simulated, and others~\cite{UCRArchive2018}. For our evaluation, we select 30 datasets from the 128 datasets in UCR, ensuring coverage across multiple categories. These datasets vary in training and test sizes, sequence lengths, and numbers of classes, making the evaluation broad and representative. Table~\ref{tab:tsc_datasets} summarizes the selected UCR datasets.

For multivariate classification, we use the UEA archive~\cite{bagnall2018ueamultivariatetimeseries}, which contains datasets with multiple input channels. We select 10 datasets from the 30 datasets in UEA, covering diverse application domains and a broad range of input dimensionalities, sequence lengths, and class counts. Table~\ref{tab:tsc_multivariate} summarizes the selected UEA datasets.

For both settings, we split the original training data into 80\% training and 20\% validation sets. Hyperparameters are selected based on validation accuracy. When multiple configurations achieve the same validation accuracy, one is randomly chosen for final training on the full training set and evaluation on the provided test set.

\begin{table}[ht]
\caption{Summary of selected UCR univariate time series classification datasets used in our experiments. The table includes dataset category, name, training and test sizes, input sequence lengths, and numbers of classes~\cite{UCRArchive2018}.}
\label{tab:tsc_datasets}
\centering
\footnotesize
\renewcommand{\arraystretch}{1.2}
\setlength{\tabcolsep}{4pt}
\begin{tabular}{llcccc}
\toprule
\textbf{Type} & \textbf{Dataset} & \textbf{Train} & \textbf{Test} & \textbf{Length} & \textbf{Classes} \\
\midrule
{\textbf{Device}}
& ACSF1 & 100 & 100 & 1460 & 10 \\
& HouseTwenty & 34 & 101 & 3000 & 2 \\
& ScreenType & 375 & 375 & 720 & 3 \\
\midrule
{\textbf{ECG}}
& ECG5000 & 500 & 4500 & 140 & 5 \\
& ECG200 & 100 & 100 & 96 & 2 \\
& ECGFiveDays & 23 & 861 & 136 & 2 \\
\midrule
{\textbf{Image}}
& Adiac & 390 & 391 & 176 & 37 \\
& FaceFour & 24 & 88 & 350 & 4 \\
& FaceAll & 560 & 1690 & 131 & 14 \\
& HandOutlines & 1000 & 370 & 2709 & 2 \\
& MiddlePhalanxTW & 399 & 154 & 80 & 6 \\
& PhalangesOutlinesCorrect & 1800 & 858 & 80 & 2 \\
& ShapesAll & 600 & 600 & 512 & 60 \\
\midrule
{\textbf{Motion}}
& Haptics & 155 & 308 & 1092 & 5 \\
& WormsTwoClass & 181 & 77 & 900 & 2 \\
& InlineSkate & 100 & 550 & 1882 & 7 \\
\midrule
{\textbf{Sensor}}
& Car & 60 & 60 & 577 & 4 \\
& Earthquakes & 322 & 139 & 512 & 2 \\
& FordA & 3601 & 1320 & 500 & 2 \\
& FordB & 3636 & 810 & 500 & 2 \\
& ItalyPowerDemand & 67 & 1029 & 24 & 2 \\
& Lightning2 & 60 & 61 & 637 & 2 \\
& StarLightCurves & 1000 & 8236 & 1024 & 3 \\
\midrule
{\textbf{Spectro}}
& Beef & 30 & 30 & 470 & 5 \\
& EthanolLevel & 504 & 500 & 1751 & 4 \\
& Wine & 57 & 54 & 234 & 2 \\
& Meat & 60 & 60 & 448 & 3 \\
& OliveOil & 30 & 30 & 570 & 4 \\
\midrule
{\textbf{Simulated/Audio}}
& ChlorineConcentration & 467 & 3840 & 166 & 3 \\
& Phoneme & 214 & 1896 & 1024 & 39 \\
\bottomrule
\end{tabular}
\end{table}

\begin{table}[ht]
\caption{Summary of selected UEA multivariate time series classification datasets used in our experiments. The table includes training and test sizes, input dimensionality, sequence length, and number of classes~\cite{bagnall2018ueamultivariatetimeseries}.}
\label{tab:tsc_multivariate}
\centering
\footnotesize
\renewcommand{\arraystretch}{1.0}
\setlength{\tabcolsep}{5pt}
\begin{tabular}{lccccc}
\toprule
\textbf{Dataset} & \textbf{Train} & \textbf{Test} & \textbf{Dim.} & \textbf{Length} & \textbf{Classes} \\
\midrule
AtrialFibrillation & 15 & 15 & 2 & 640 & 3 \\
Cricket & 108 & 72 & 6 & 1197 & 12 \\
DuckDuckGeese & 60 & 40 & 1345 & 270 & 5 \\
ERing & 30 & 30 & 4 & 65 & 6 \\
EthanolConcentration & 261 & 263 & 3 & 1751 & 4 \\
LSST & 2459 & 2466 & 6 & 36 & 14 \\
Libras & 180 & 180 & 2 & 45 & 15 \\
FaceDetection & 5890 & 3524 & 144 & 62 & 2 \\
FingerMovements & 316 & 100 & 28 & 50 & 2 \\
MotorImagery & 278 & 100 & 64 & 3000 & 2 \\
\bottomrule
\end{tabular}
\end{table}

\clearpage
\end{document}